\documentclass{article}

\usepackage{graphicx}
\usepackage{amsmath}
\usepackage{amssymb}
\usepackage{multirow}
\usepackage{colortbl}
\usepackage{caption}
\usepackage{wrapfig}
\usepackage{booktabs} % 推荐使用，用于绘制专业的“三线表”
\usepackage{multirow}
\usepackage{graphicx}
\usepackage{caption}
\usepackage[table]{xcolor}
\usepackage{hyperref}

\usepackage[final]{corl_2026} % Uncomment for the camera-ready ``final'' version.

\title{MosaicIMU: Composing Carrier Experts for Generalizable Neural Inertial Odometry}

\author{
\textbf{Junye Zou$^{1,*}$,\quad Huiyi Yan$^{2,*}$,\quad Xinning Xu$^{3}$,\quad Xiaolei Li$^{4}$,}\\
\textbf{Pengkun Zhou$^{1}$,\quad Jinhui Zhang$^{5}$,\quad Ziyang Meng$^{1,\dagger}$}\\[1.2ex]
\normalfont
$^{1}$State Key Laboratory of Precision Measurement Technology and Instruments,\\
Department of Precision Instrument, Tsinghua University,\\
$^{2}$Xi'an Jiaotong University,\quad
$^{3}$Beijing University of Chemical Technology\\
$^{4}$Beijing Information Science and Technology University,\\
$^{5}$State Key Laboratory of Autonomous Intelligent Unmanned Systems,\\
School of Automation, Beijing Institute of Technology\\[1.0ex]
$^{*}$Equal contribution,\qquad
$^{\dagger}$Corresponding author\\[1.0ex]
}

\begin{document}
\maketitle

\begingroup
\renewcommand{\thefootnote}{}
\footnotetext{
E-mail: \texttt{zoujy21@mails.tsinghua.edu.cn} (Junye Zou); \texttt{ziyangmeng@mail.tsinghua.edu.cn} (Ziyang Meng).
}
\addtocounter{footnote}{-1}
\endgroup

%===============================================================================

\begin{abstract}
    Robust inertial odometry is essential for various carriers when external sensing is unreliable. Learning-based methods reduce integration drift by capturing local motion priors, but these methods often remain tied to a particular carrier, limiting generalization across heterogeneous platforms. We present \textbf{MosaicIMU}, a carrier-conditioned Mixture-of-Experts (MoE) pretraining-and-adaptation framework for generalizable neural inertial odometry. 
MosaicIMU uses a prototype-based router to compose carrier-specific expert features, decodes local velocity and uncertainty constraints, and integrates them with a history-aware EKF. For unseen domain adaptation, it freezes the pretrained base model and learns a new lightweight expert residual branch. 
For edge-deployment, it further reuses the router to select informative online samples for efficient incremental updates. Experiments show that MosaicIMU outperforms the strongest learning-based baseline, reducing average ATE and RTE-10s by 40\% and 34\%, respectively. 
These results highlight that MosaicIMU provides a scalable pretraining-to-deployment paradigm for generalizable and adaptive neural inertial odometry.

\end{abstract}


\begin{quote}
\textbf{Code:} \url{https://github.com/zoujy02/MosaicIMU}
\end{quote}

\section{Introduction}

Inertial Measurement Units (IMUs) are widely used for motion estimation on vehicles, pedestrians, quadruped robots, and drones~\citep{ahmad2013reviews}.
They are lightweight, low-cost, and robust in visually degraded scenarios, making them attractive when cameras or LiDARs become unreliable~\citep{zhao2021super}. 
However, inertial odometry is inherently prone to drift, as small errors from bias, vibration, temperature variation, and sensor noise are continuously accumulated through integration~\citep{kok2017using}.

Learning-based inertial odometry mitigates such drift by learning local motion priors from IMU sequences, 
which capture carrier-dependent relationships between inertial patterns and local motion estimates.
Existing methods estimate displacement, velocity, uncertainty, or motion constraints from inertial windows~\citep{chen2018ionet,chen2019motiontransformer,liu2020tlio}, achieving promising results on vehicles~\citep{brossard2020ai,zou2025mtnet}, pedestrians~\citep{liu2020tlio,herath2020ronin}, quadruped robots~\citep{buchanan2022deep}, and drones~\citep{cioffi2023learned,qiu2025airio}.
However, most models are designed for a specific carrier and motion regime. 
When transferred to another carrier, their learned priors may fail.

A natural expectation is to construct a pretrained foundation model that can be used for heterogeneous carriers. 
Recent foundation-style IMU models, like TartanIMU, show that large-scale pretraining can learn shared inertial representations across humans, vehicles, robots, and aerial platforms~\citep{zhao2025tartan}. 
However, the existing multi-carrier design usually relies on a shared backbone with separate carrier-specific heads. It treats carrier identity as a rigid one-hot condition, and confines cross-carrier knowledge transfer mostly to the shared backbone. 
In practice, carrier dynamics are not strictly decoupled: different carriers have their own motion characteristics, but they also share common inertial patterns.
This raises a key question: \emph{how can we preserve carrier-specific dynamic priors while allowing sample-adaptive feature composition across carriers?}

To this end, we propose \textbf{MosaicIMU}, a carrier-conditioned pretraining-and-adaptation framework for generalizable neural inertial odometry across heterogeneous carriers. The name reflects our view that carrier-dependent motion priors should be represented not by a fixed carrier-specific head, but by a sample-adaptive mosaic of carrier-expert features. Our contributions are summarized as follows:

\begin{itemize}

    \item We propose MosaicIMU, a carrier-conditioned Mixture-of-Experts (MoE) framework that achieves sample-adaptive, feature-level expert composition, and integrates local velocity constraints with a history-aware EKF for stable state estimation.
    
    \item For better performance on unseen domains, such as new IMU mounting conditions, new sensor configurations, and new motion patterns, we freeze the pretrained base model and introduce a new lightweight expert residual branch for efficient adaptation.

    \item For resource-constrained online deployment, we further develop router-guided sample selection, which maintains a compact online buffer of informative samples for efficient incremental updates. 
    
    \item Extensive experiments demonstrate strong cross-carrier generalization, reducing the average ATE/RTE-10s by 40\%/34\% compared with the strongest learning-based baseline. For unseen domains, residual adaptation achieves 1.49 m ATE on the TLIO dataset, while router-guided online adaptation reduces ATE from 42.60 m to 7.60 m on self-collected sequences. These results establish that MosaicIMU provides a scalable path toward generalizable and adaptive neural inertial odometry.
\end{itemize}

\section{Related Work}\label{sec:Relate}

\paragraph{Learning-based inertial odometry.}
Classical inertial odometry integrates accelerometer and gyroscope measurements to estimate motion, but small errors from sensor bias, noise, vibration, and gravity compensation rapidly accumulate and cause drift. 
Learning-based methods alleviate this issue by inferring local motion constraints from IMU sequences. 
Early works such as IONet~\citep{chen2018ionet}, RoNIN~\citep{herath2020ronin}, and MotionTransformer~\citep{chen2019motiontransformer} regress relative displacement or velocity from inertial windows using deep sequence models. 
Subsequent studies improve temporal modeling with recurrent, convolutional, or attention-based architectures~\citep{esfahani2019aboldeepio,feigl2019bidirectional,li2022calib,liu2023smartphone,rao2022ctin,zou2025mtnet}, and enhance robustness through rotation-equivariant learning, frame-aligned representations, or physics-guided inertial features~\citep{jayanth2025eqnio,qiu2025airio}. 
These methods demonstrate the effectiveness of learning-based inertial odometry, but most are designed for a specific carrier and motion regime.

\paragraph{Carrier-specific and multi-carrier inertial learning.}
Learned inertial priors are closely coupled with carrier dynamics. 
Pedestrian methods exploit human motion regularities~\citep{herath2020ronin,liu2020tlio}, vehicle methods often rely on planar or non-holonomic assumptions~\citep{brossard2020ai,zhou2024learning}, legged-robot methods focus on contact-rich locomotion and bias estimation~\citep{buchanan2022deep,buchanan2022learning}, and aerial methods target highly dynamic UAV motion~\citep{cioffi2023learned}. 
Such assumptions work well within each carrier category, but limit generalization across heterogeneous carriers.
Recent foundation-style IMU models, such as TartanIMU, pretrain on large-scale heterogeneous data and use carrier-specific output heads to estimate velocity~\citep{zhao2025tartan}. 
We follow this pipeline but replace the fixed output partition with feature-level adaptive fusion of different carriers, including vehicles, pedestrians, quadruped robots, and drones.

\paragraph{Estimator-compatible learning-based inertial constraints.}
Another key issue is how learned predictions are represented and fused with a state estimator. 
Prior works suggest that network outputs should serve as estimator-compatible constraints rather than isolated regression objectives. 
TLIO fuses learned displacement and uncertainty in a tightly coupled EKF~\citep{liu2020tlio}, RNIN-VIO incorporates neural inertial estimates into visual-inertial odometry~\citep{chen2021rnin}, and legged inertial odometry integrates learned IMU displacement into factor-graph estimation~\citep{buchanan2022learning}. 
Similarly, we predict dense velocity sequences in a local gravity-aligned frame, avoiding introducing global-yaw observations while maintaining a stable local reference within each window. 
The predicted sequence is then aggregated into a segment-level velocity observation for a history-aware EKF, improving robustness to frame-wise prediction noise.

\begin{figure*}[t]
  \centering
\includegraphics[width=\linewidth]{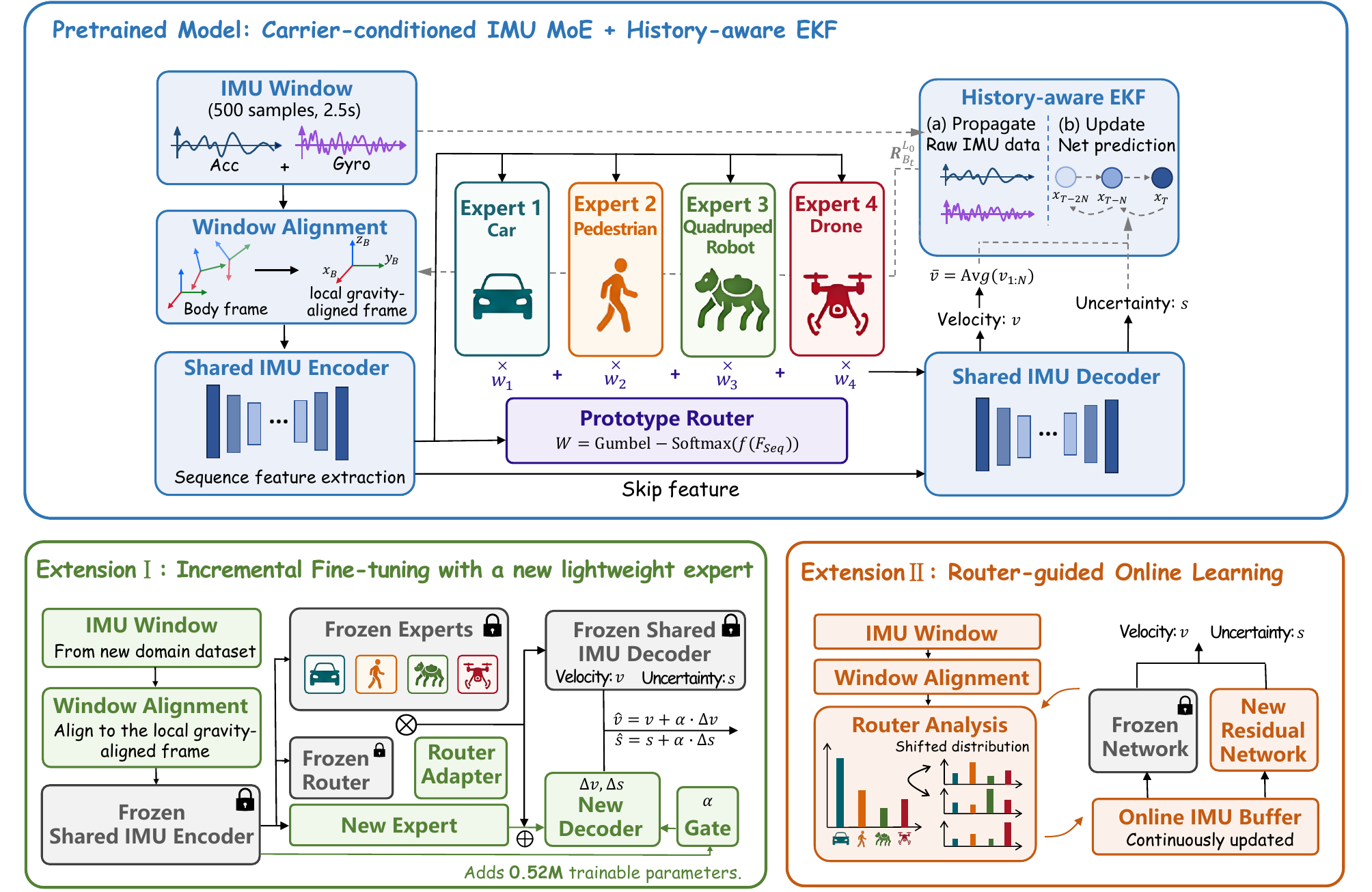}
  \caption{\textbf{Overall architecture of MosaicIMU.} }
  \label{fig:framework}
  \vspace{-0.3cm}
\end{figure*}
\vspace{-0.2cm}
\section{Method}
\label{sec:method}
% \subsection{Overview}
% \label{sec:pipeline}
As shown in Figure~\ref{fig:framework}, MosaicIMU is organized as one pretrained base model and two adaptation extensions. 
We first train a multi-carrier pretrained MosaicIMU base model, which learns generalizable inertial representations across heterogeneous carriers and integrates the resulting velocity observations with a history-aware EKF. 
Extension I targets offline adaptation to unseen domains where a target dataset is available. 
In such a case, we freeze the pretrained base model and update only a lightweight expert residual branch for efficient fine-tuning. 
Extension II targets online adaptation when target data arrive continuously during online deployment, and the edge computation resource is limited. 
We then introduce router-guided online sample selection to maintain a compact online buffer of informative samples for efficient incremental updates.
\subsection{Carrier-Conditioned Mixture-of-Experts Network}
\label{sec:moe}

\paragraph{Local gravity-aligned IMU representation.}
Inspired by TLIO~\citep{liu2020tlio}, we express each IMU window in a local gravity-aligned frame $L_0$, whose vertical axis is aligned with gravity and whose yaw matches the body heading at the start of the window.
Let $\mathbf{R}_{B_t}^{W}$ denote the rotation from the instantaneous body frame $B_t$ to the world frame $W$, whose vertical axis is also aligned with gravity.
The initial body yaw defines $\mathbf{R}_{z,0}$, the rotation from $L_0$ to $W$, giving $\mathbf{R}_{B_t}^{L_0} = \mathbf{R}_{z,0}^{\top}\mathbf{R}_{B_t}^{W}$.
We transform the accelerometer and gyroscope measurements into $L_0$ and express the ground-truth velocity in the same frame:
\begin{equation}
    \mathbf{x}^{L_0}_{t}
    =
    [\mathbf{R}_{B_t}^{L_0}\mathbf{a}^{B_t}_{t};
    \mathbf{R}_{B_t}^{L_0}\boldsymbol{\omega}^{B_t}_{t}],
    \qquad
    \mathbf{v}^{L_0}_{t}
    =
    \mathbf{R}_{z,0}^{\top}\mathbf{v}^{W}_{t}.
\end{equation}
This representation provides a fixed reference within each window while remaining invariant to the choice of global yaw reference.

\paragraph{Carrier-conditioned feature fusion.}
Given the aligned sequence $\mathbf{X}_{1:T}^{L_0}=\{ \mathbf{x}^{L_0}_{t} \}_{1:T}$, a shared encoder $E_{\theta}$ extracts temporal and skip features:
\begin{equation}
    \mathbf{F},\mathbf{u}_1,\mathbf{u}_2
    =
    E_{\theta}(\mathbf{X}_{1:T}^{L_0}),
\end{equation}
where $\mathbf{F}$ is the encoded inertial feature, and $\mathbf{u}_1,\mathbf{u}_2$ are multi-scale skip features for decoding. 
The router $g_{\phi}$ maps $\mathbf{F}$ to $\mathbf{z}=g_{\phi}(\mathbf{F})$, and compares it with $K$ learnable carrier prototypes $\{\mathbf{p}_k\}_{k=1}^{K}$:
\begin{equation}
    e_k
    =
    \gamma
    \frac{\mathbf{z}^{\top}\mathbf{p}_k}
    {\tau\|\mathbf{z}\|\|\mathbf{p}_k\|},
    \qquad
    \{w_k\}_{1:K}
    =
    \operatorname{Gumbel-Softmax}([e_1,\ldots,e_K]),
\end{equation}
where $e_k$ is the similarity logit, $w_k$ is the routing weight, $\gamma$ is a learnable scale, and $\tau$ is the temperature. 
Gumbel-Softmax is used to encourage the exploration of the feature composition. 
In this work, $K$ is specified to 4, which represents vehicle, pedestrian, quadruped robot, and drone, respectively.

Under the MoE architecture, each expert $G_k$ produces a carrier-specific feature. 
Instead of using separate output heads, we use the routing weights to compose expert features at the feature level before a shared decoder $D_{\theta}$:
\begin{equation}
    \mathbf{F}_{moe}
    =
    \sum_{k=1}^{K} w_k G_k(\mathbf{F}),
    \qquad
    \hat{\mathbf{v}}^{L_0}_{1:T},\hat{\mathbf{s}}_{1:T}
    =
    D_{\theta}(\mathbf{F}_{moe},\mathbf{u}_1,\mathbf{u}_2),
\end{equation}
where $\mathbf{F}_{moe}$ is the fused feature, $\hat{\mathbf{v}}^{L_0}_{1:T}$ is the dense velocity prediction in the local gravity-aligned frame $L_0$, and $\hat{\mathbf{s}}_{1:T}$ is the predicted velocity variance. 
This enables sample-adaptive composition of carrier-specific motion priors.

\paragraph{Training objective.}
We train the base model to predict velocities and their uncertainty, while encouraging distinct carrier prototypes and carrier-aware expert composition. The overall loss function is:
\begin{equation}
\label{eq:loss}
    \mathcal{L}_{base}
    =
    \mathcal{L}_{NLL}
    +
    \mathcal{L}_{vel}
    +
    \lambda_{orth}\mathcal{L}_{orth}
    +
    \lambda_{car}\mathcal{L}_{car},
\end{equation}
where $\lambda_{orth}$ and $\lambda_{car}$ balance the two regularization terms.

The heteroscedastic Gaussian negative log-likelihood jointly supervises velocity and uncertainty prediction, while an auxiliary squared-error loss provides direct velocity supervision without uncertainty weighting:
\begin{equation}
    \mathcal{L}_{NLL}
    =
    \frac{1}{T}
    \sum_{t=1}^{T}
    \left[
    \frac{1}{2}
    (\mathbf{v}^{L_0}_{t}-\hat{\mathbf{v}}_{t})^{\top}
    \hat{\mathbf{S}}_{t}^{-1}
    (\mathbf{v}^{L_0}_{t}-\hat{\mathbf{v}}_{t})
    +
    \frac{1}{2}\log |\hat{\mathbf{S}}_{t}|
    \right],
    \quad
    \mathcal{L}_{vel}
    =
    \frac{1}{T}
    \sum_{t=1}^{T}
    \|\hat{\mathbf{v}}_{t}-\mathbf{v}^{L_0}_{t}\|_2^2,
\end{equation}
where $\hat{\mathbf{S}}_{t}=\operatorname{diag}(\hat{\mathbf{s}}_{t})$ is the predicted velocity covariance, and $\hat{\mathbf{v}}_{t}$ is expressed in $L_0$.

To discourage prototype collapse and promote distinct routing references, we penalize the absolute cosine similarity between carrier prototypes:
\begin{equation}
    \mathcal{L}_{orth}
    =
    \frac{2}{K(K-1)}
    \sum_{1\le i<j\le K}
    \left|
    \bar{\mathbf{p}}_i^{\top}\bar{\mathbf{p}}_j
    \right|,
    \qquad
    \bar{\mathbf{p}}_i
    =
    \frac{\mathbf{p}_i}{\|\mathbf{p}_i\|_2}.
\end{equation}

To further associate experts with carrier categories while retaining sample-adaptive composition, we regularize the routing distribution $\mathbf{w}$ towards a carrier-specific prior $\mathbf{q}_c=[q_{c,1},\ldots,q_{c,K}]^{\top}$:
\begin{equation}
    \mathcal{L}_{car}
    =
    D_{\mathrm{KL}}(\mathbf{w}\|\mathbf{q}_c)
    =
    \sum_{k=1}^{K}
    w_k
    \left[
    \log(w_k+\epsilon)
    -
    \log(q_{c,k}+\epsilon)
    \right],
\end{equation}
where $c$ is the carrier label, $q_{c,k}\geq 0$, $\sum_{k=1}^{K}q_{c,k}=1$, and $\epsilon$ is a small constant for numerical stability.

\subsection{Segment-Level Velocity Constraints for History-Aware EKF}
\label{sec:ekf}

Dense velocity predictions provide rich local constraints, but using every frame as an EKF measurement is sensitive to regression noise.
We therefore aggregate the dense predictions within each window into a single segment-level average-velocity observation between its two endpoint clones.
The filter follows a cloned-pose error-state formulation~\citep{roumeliotis2002stochastic,mourikis2007multi}.
Unlike displacement-based neural inertial filters~\citep{liu2020tlio,cioffi2023learned}, MosaicIMU learns dense local velocities and converts them into velocity-consistency constraints between historical pose pairs.

The filter state is
$
\mathbf{x}=[\mathbf{R}^{W}_{B},\mathbf{v}^{W},\mathbf{p}^{W},\mathbf{b}_{g},\mathbf{b}_{a}],
$
where $\mathbf{R}^{W}_{B}$, $\mathbf{v}^{W}$, and $\mathbf{p}^{W}$ are the IMU orientation, velocity, and position in the world frame, and $\mathbf{b}_{g},\mathbf{b}_{a}$ are gyroscope and accelerometer biases.
To apply segment-level updates, the filter maintains $N$ historical pose clones,
$
\mathcal{C}=\{(\mathbf{R}^{W}_{B_n},\mathbf{p}^{W}_{n})\}_{n=1}^{N}.
$
Raw IMU measurements propagate the current state, whose pose is cloned at regular intervals.
Once sufficient history and a complete prediction window are available, each new clone is paired with a historical clone a fixed number of cloning intervals earlier for one update, after which the oldest clone is removed.

For a prediction window associated with clones $i$ and $j$, let $\mathcal{I}_{ij}$ contain its $T$ prediction frames.
We aggregate the network outputs as
\begin{equation}
    \mathbf{z}_{ij}
    = \frac{1}{T}\sum_{t\in\mathcal{I}_{ij}}\hat{\mathbf{v}}^{L_i}_{t},
    \qquad
    \bar{\mathbf{N}}_{ij}
    = \operatorname{diag}\left(
    \frac{1}{T^2}\sum_{t\in\mathcal{I}_{ij}}\hat{\mathbf{s}}^{L_i}_{t}
    \right),
\end{equation}
where $\mathbf{z}_{ij}$ is the neural average-velocity observation and $\bar{\mathbf{N}}_{ij}$ is its approximate diagonal covariance.
The filter prediction is the average velocity implied by the cloned poses, expressed in the same local gravity-aligned frame $L_i$:
\begin{equation}
    h_{ij}(\mathcal{C})
    = \mathbf{R}_{z,i}^{\top}
    \frac{\mathbf{p}^{W}_{j}-\mathbf{p}^{W}_{i}}{\Delta t_{ij}},
    \qquad
    \mathbf{r}_{ij}
    = \mathbf{z}_{ij}-h_{ij}(\mathcal{C}),
\end{equation}
where $\Delta t_{ij}$ is the time interval between the two clones, $\mathbf{R}_{z,i}$ is the yaw-only rotation from $L_i$ to the world frame, determined by the initial clone's heading, and $\mathbf{r}_{ij}$ is the EKF residual.
This residual enforces consistency between the learned local velocity and the historical trajectory without introducing an absolute global-yaw observation.
Detailed propagation and update derivations are provided in Appendix~\ref{app:ekf}.

\subsection{Incremental Fine-tuning with a New Lightweight Expert}
\label{sec:incremental}

Although MosaicIMU learns transferable inertial priors, unseen domains may introduce distribution shifts that require adaptation.
In Extension I, we adapt the model using available target-domain data by freezing the pretrained base model and training only a router adapter and a gated residual branch.

The router adapter adjusts how pretrained experts are combined for the target domain.
Given the encoded feature $\mathbf{F}$, it predicts an additive correction to the frozen base router logits $\boldsymbol{\ell}_{old}$:
\begin{equation} 
\label{eq:router_adapter}   
    \Delta\boldsymbol{\ell}=A_{\eta}(\mathbf{F}),
    \qquad
    \mathbf{w}'
    =
    \operatorname{Gumbel-Softmax}
    \left(
    \boldsymbol{\ell}_{old}+\beta\Delta\boldsymbol{\ell}
    \right),
\end{equation}
where $\Delta\boldsymbol{\ell}$ is the correction, $\mathbf{w}'$ is the calibrated routing distribution, and $\beta$ controls the correction strength. 
The calibrated prediction is obtained from the frozen experts $G_k^{base}$ and decoder $D_{\theta}^{base}$:
\begin{equation}
    \mathbf{F}_{old}
    =
    \sum_{k=1}^{K}w'_k G_k^{base}(\mathbf{F}),
    \qquad
    (\hat{\mathbf{v}}_{o},\hat{\mathbf{s}}_{o})
    =
    D_{\theta}^{base}(\mathbf{F}_{old},\mathbf{u}_1,\mathbf{u}_2),
\end{equation}
where $\hat{\mathbf{v}}_{o}$ and $\hat{\mathbf{s}}_{o}$ are the base velocity and variance predictions after routing weight calibration.

A lightweight expert $G_{\psi}^{new}$ and decoder $D_{\psi}^{new}$ then predict target-domain residuals:
\begin{equation}
    \mathbf{F}_{new}=G_{\psi}^{new}(\mathbf{F}),
    \qquad
    \Delta\mathbf{v},\Delta\mathbf{s}
    =
    D_{\psi}^{new}([\mathbf{F}_{old},\mathbf{F}_{new}],\mathbf{u}_1,\mathbf{u}_2),
\end{equation}
where  $\Delta\mathbf{v},\Delta\mathbf{s}$ are velocity and variance residuals. 
A gate $q_{\psi}$ controls their magnitude:
\begin{equation}
    \alpha=\operatorname{sigmoid}(q_{\psi}(\mathbf{F})),
    \qquad
    \hat{\mathbf{v}}=\hat{\mathbf{v}}_{o}+\alpha\Delta\mathbf{v},
    \quad
    \hat{\mathbf{s}}=\hat{\mathbf{s}}_{o}+\alpha\operatorname{softplus}(\Delta\mathbf{s}),
\end{equation}
where $\alpha$ adaptively scales the correction. We use the same loss function as in Eq. (\ref{eq:loss}), except that the term $\mathcal{L}_{orth}$ is omitted.

\subsection{Router-Guided Online Adaptation}
\label{sec:online}

Unlike Extension I, which utilizes an available offline dataset, Extension II targets real-time online adaptation where IMU windows arrive sequentially and the edge computation resource is limited. We reuse the router as an indicator of carrier-distribution mismatch to reduce the computational burden of online updates. 
Let $\mathcal{S}_{online}=\{\mathbf{X}_i\}$ denote the recent IMU windows collected from the online stream. For an expected carrier $c$, the mismatch score $m$ is
\begin{equation}
    m(\mathbf{X})=1-\tilde{w}_c(\mathbf{X}),
\end{equation}
where $\tilde{w}_c(\mathbf{X})$ is the routing probability of carrier $c$ from the calibrated router. 
We maintain a compact buffer $\mathcal{B}$ by selecting high-mismatch samples:
\begin{equation}
    \mathcal{B}
    \leftarrow
    \operatorname{TopM}_{\mathbf{X}\in\mathcal{S}_{online}}
    m(\mathbf{X}).
\end{equation}
Samples in the buffer are used to update the same lightweight components in Section~\ref{sec:incremental}, while the base MosaicIMU parameters remain frozen.

\begin{figure*}[!t] % 或者使用 [p] 如果它们占满了整整一页
    \centering
    
    % ==========================================
    % 1. 顶部的图片
    % ==========================================
    \includegraphics[width=0.95\linewidth, height=0.65\linewidth]{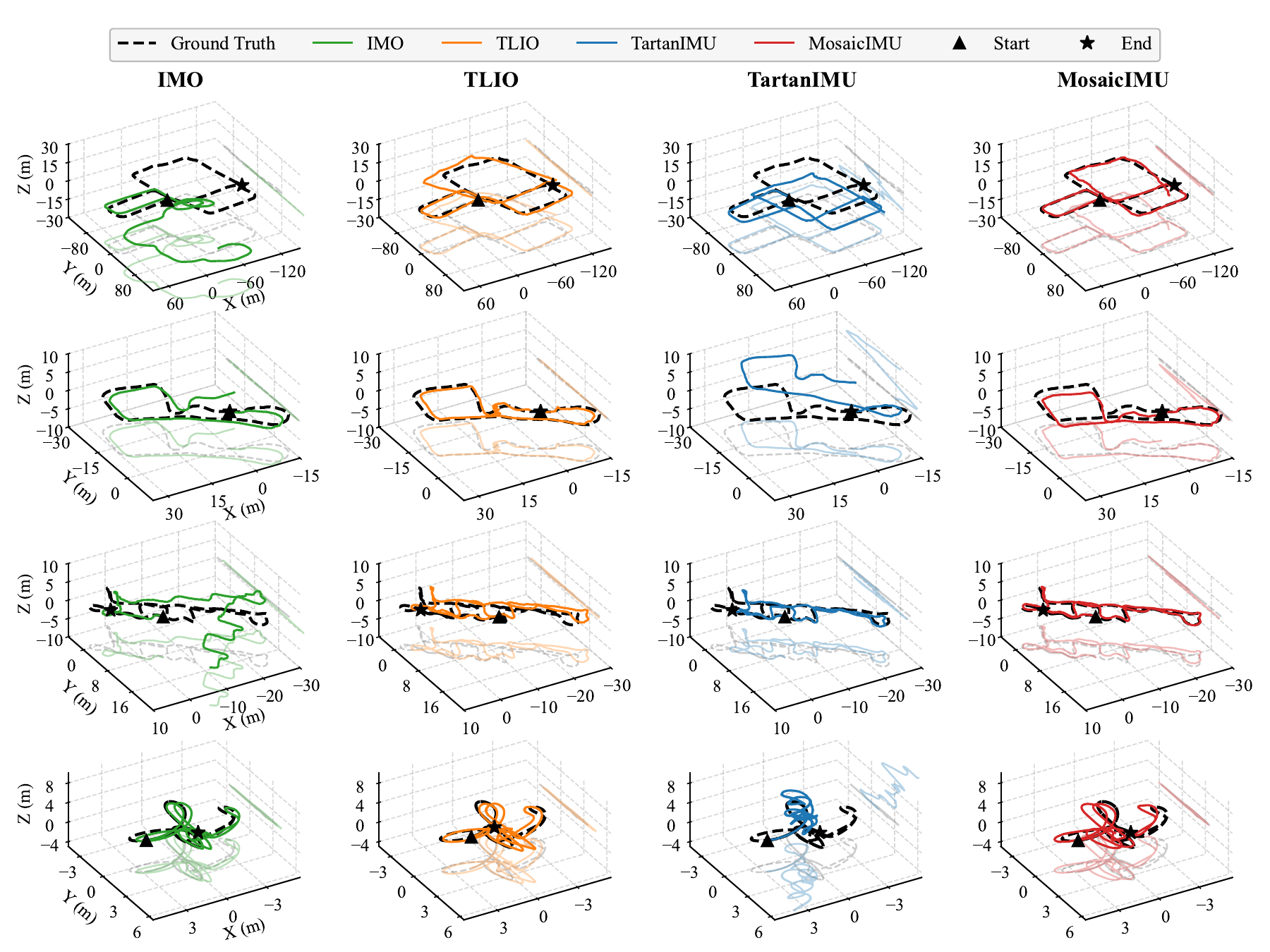}
    \caption{\textbf{Comparison of estimated trajectories with ground truth across heterogeneous IMU carriers.} From top to bottom, the four rows correspond to vehicle, quadruped robot, pedestrian, and drone sequences, respectively.}
    \label{fig:pretrain_model}

    % ==========================================
    % 2. 中间的大表格
    % ==========================================
    \captionsetup{type=table}
\caption{\textbf{Generalization evaluation on held-out test sequences.} Unit: m.}
\label{tab:generalization}
\renewcommand{\arraystretch}{1.2}
\setlength{\arrayrulewidth}{1pt}
\resizebox{0.8\linewidth}{!}{%
    \begin{tabular}{c|cc|cc|cc|cc} 
    \hline
    \multirow{2}{*}{\textbf{Carrier}} 
    & \multicolumn{2}{c|}{\textbf{IMO}} 
    & \multicolumn{2}{c|}{\textbf{TLIO}} 
    & \multicolumn{2}{c|}{\textbf{TartanIMU}} 
    & \multicolumn{2}{c}{\textbf{MosaicIMU}} \\
    & ATE $\downarrow$ & RTE-10s $\downarrow$ 
    & ATE $\downarrow$ & RTE-10s $\downarrow$ 
    & ATE $\downarrow$ & RTE-10s $\downarrow$  
    & ATE $\downarrow$ & RTE-10s $\downarrow$ \\ 
    \hline

    Vehicle    
    & 156.02 & 13.11 
    & 20.87 & 3.68 
    & 25.44 & 2.77 
    & \cellcolor[rgb]{1,1,0.741}\textbf{10.56} \textcolor[rgb]{0,0.502,0}{49\%$\downarrow$} 
    & \cellcolor[rgb]{1,1,0.741}\textbf{2.17} \textcolor[rgb]{0,0.502,0}{22\%$\downarrow$} \\

    Quadruped 
    & 3.49 & 1.22 
    & \textbf{1.74} & 0.79 
    & 3.44 & 1.17 
    & \cellcolor[rgb]{1,1,0.741}{2.01} \textcolor[rgb]{1,0,0}{16\%$\uparrow$} 
    & \cellcolor[rgb]{1,1,0.741}\textbf{0.63} \textcolor[rgb]{0,0.502,0}{20\%$\downarrow$} \\

    Pedestrian  
    & 11.91 & 6.21 
    & 3.83 & 1.95 
    & 10.12 & 2.39 
    & \cellcolor[rgb]{1,1,0.741}\textbf{2.85} \textcolor[rgb]{0,0.502,0}{26\%$\downarrow$} 
    & \cellcolor[rgb]{1,1,0.741}\textbf{1.28} \textcolor[rgb]{0,0.502,0}{34\%$\downarrow$} \\

    Drone      
    & 1.29 & 1.17 
    & 1.37 & 1.48 
    & 8.74 & 6.08 
    & \cellcolor[rgb]{1,1,0.741}\textbf{1.17} \textcolor[rgb]{0,0.502,0}{9\%$\downarrow$} 
    & \cellcolor[rgb]{1,1,0.741}\textbf{1.14} \textcolor[rgb]{0,0.502,0}{3\%$\downarrow$} \\ 

    \hline

    Average    
    & 43.18 & 5.43 
    & 6.95 & 1.98 
    & 11.94 & 3.10 
    & \cellcolor[rgb]{1,1,0.741}\textbf{4.15} \textcolor[rgb]{0,0.502,0}{40\%$\downarrow$} 
    & \cellcolor[rgb]{1,1,0.741}\textbf{1.31} \textcolor[rgb]{0,0.502,0}{34\%$\downarrow$} \\

    \hline
    \end{tabular}%
}

\end{figure*}

\section{Experimental Results}
\label{sec:result}
We evaluate MosaicIMU from three aspects: multi-carrier generalization, lightweight adaptation to unseen domains, and router-guided online learning. We report Absolute Trajectory Error (ATE) for global trajectory accuracy, Relative Trajectory Error over 10 seconds (RTE-10s) for local drift, and $z$-Drift, defined as ATE computed along the $z$ axis, for vertical positioning accuracy.
Implementation details, datasets, and baselines are provided in Appendix \ref{app:dataset_details} and Appendix \ref{app:baseline}.

\subsection{Generalization of the Pretrained MosaicIMU}
\label{sec:exp_generalization}

We first compare MosaicIMU with IMO~\citep{cioffi2023learned}, TLIO~\citep{liu2020tlio}, and TartanIMU~\citep{zhao2025tartan}. 
For IMO and TLIO, we train one model per carrier category, yielding four carrier-specific models for each baseline,  while TartanIMU and MosaicIMU are trained on the same multi-carrier pretraining datasets. All methods are evaluated on identical held-out trajectories.

As shown in Table~\ref{tab:generalization} and Figure~\ref{fig:pretrain_model}, MosaicIMU achieves the best average performance and the lowest ATE across all the baselines. 
Compared with TartanIMU, which uses fixed carrier-specific heads, our method reduces the average ATE from $11.94$ m to $4.15$ m and RTE-10s from $3.10$ m to $1.31$ m. Meanwhile, it also outperforms other learning-based inertial odometry baselines.
The gains are especially clear on vehicle and pedestrian sequences, where heterogeneous motion patterns are difficult to capture with a fixed output partition. 
These results suggest that sample-adaptive feature-level composition provides a more flexible inertial prior than either carrier-specific training or fixed-head multi-carrier training.

\begin{wraptable}{r}{0.5\textwidth}
\vspace{-0.9cm}
    \centering
    \caption{\textbf{Ablation on prediction frames.} Results are averaged over all test carriers. Unit: m.}
    \label{tab:frame_ablation}
    \renewcommand{\arraystretch}{1.3}
    \resizebox{0.95\linewidth}{!}{
        \begin{tabular}{cccc} 
            \hline
            Prediction Frame & ATE $\downarrow$ & RTE-10s $\downarrow$ & $z$-Drift $\downarrow$ \\
            \hline
            Body $B_t$       & 8.00 & 1.93 & 1.98 \\
            Global $W$       & 10.40 & 2.74 & 1.09 \\
            Local-GA $L_0$ & \textbf{4.15} & \textbf{1.31} & \textbf{0.73} \\
            \hline
        \end{tabular}
    }
\vspace{-0.3cm}
\end{wraptable}

We further ablate the velocity prediction frame in Table~\ref{tab:frame_ablation}. 
Global-frame prediction depends on the recursively estimated attitude, so rotation errors may be injected into the velocity observation and increase trajectory error. 
Body-frame prediction avoids this global-frame dependence, but the body frame moves and rotates with the carrier each time, making dense velocity predictions less stable.  
In contrast, the local gravity-aligned frame decouples the velocity observation from global yaw while preserving a gravity-consistent local reference, leading to the lowest ATE, RTE-10s, and $z$-Drift.

\begin{wraptable}{r}{0.48\textwidth}

    \centering
    \caption{\textbf{Ablation on expert design.} Results are averaged over all test carriers. Unit: m.}
    \label{tab:moe_ablation}
    \renewcommand{\arraystretch}{1.2}
    \resizebox{0.8\linewidth}{!}{
        \begin{tabular}{ccc} 
            \hline
            Model Variant & ATE $\downarrow$ & RTE-10s $\downarrow$ \\
            \hline
            Single-carrier expert    & 5.47 & 2.09 \\
            Mixture single expert    & 5.53 & 1.74 \\
            MosaicIMU            & \textbf{4.15} & \textbf{1.31} \\
            \hline
        \end{tabular}
    }
    \vspace{-0.3cm}
\end{wraptable}

Table~\ref{tab:moe_ablation} further analyzes the design of the expert architecture. 
The \emph{Single-carrier expert} variant trains four independent single-expert models, one for each carrier, without router-based knowledge sharing. 
The \emph{Mixture single expert} variant trains one expert using data from all four carriers, also without a router. 
The former benefits from carrier-specific training but cannot transfer useful priors across carriers, while the latter uses all training data but must represent heterogeneous dynamics with a single shared expert.
MosaicIMU achieves the lowest ATE and RTE-10s, showing that multi-carrier inertial learning requires both specialized carrier priors and sample-adaptive feature-level composition. We include detailed ablations in Appendix~\ref{app:ablation_pretrain} and further analyze the  router behavior in Appendix~\ref{app:router_analysis}.

\subsection{Incremental Fine-tuning with a New Lightweight Expert on Unseen Domains}
\label{sec:exp_finetune}

Although multi-carrier pretraining improves generalization, unseen datasets can still introduce sensor noise or motion-distribution shifts. 
We evaluate adaptation in a head-mounted IMU scenario from the TLIO dataset, which is excluded from pretraining. 
During fine-tuning, the base MosaicIMU model is frozen, and only the router adapter and gated residual branch are updated. 
For comparison, we adapt TartanIMU with LoRA~\citep{hu2021lora} under a similar pipeline.

\begin{figure}[t]
    \centering
    \includegraphics[width=\textwidth]{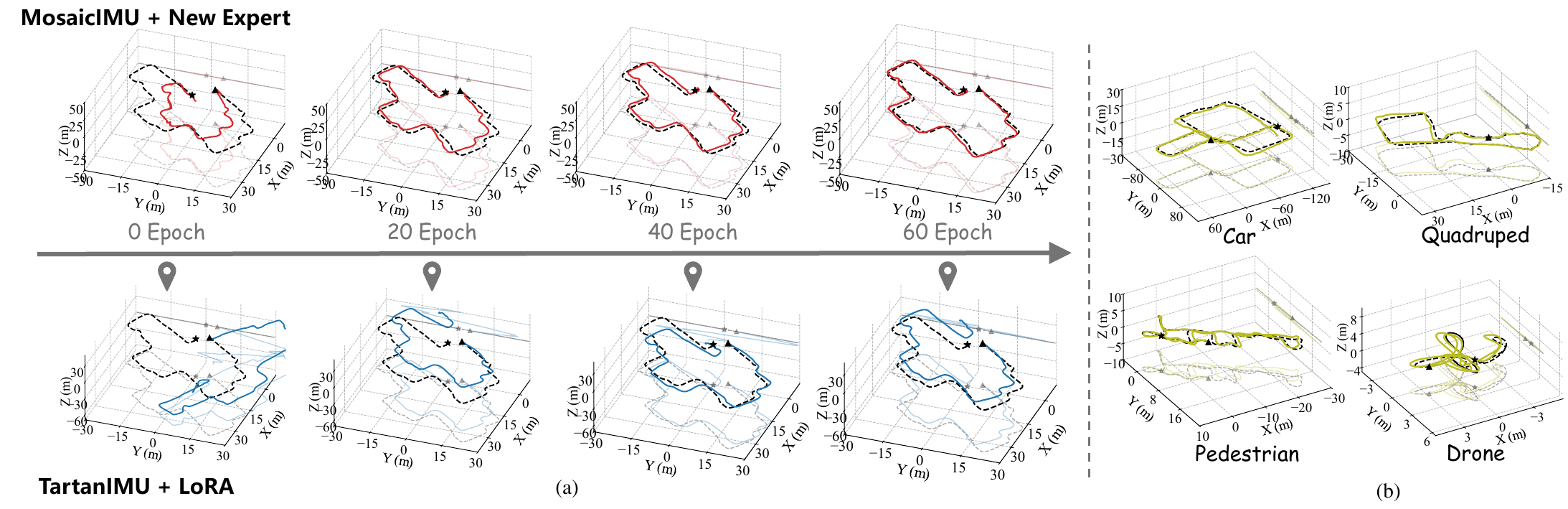}
    \caption{\textbf{Fine-tuning adaptation and forgetting analysis.} 
    (a) Trajectory evolution on the unseen TLIO dataset, where MosaicIMU with the new expert residual branch converges faster than TartanIMU with LoRA. 
    (b) Source-domain retention after target-domain fine-tuning, showing that the side-branch design preserves the pretrained multi-carrier knowledge.}
    \label{fig:finetune}
\end{figure}

Table~\ref{tab:finetune} and Figure~\ref{fig:finetune}(a) show the effectiveness of the proposed lightweight adaptation branch. 
Before fine-tuning, MosaicIMU already achieves lower error than TartanIMU+LoRA, indicating stronger zero-shot generalization from multi-carrier pretraining. 
With only $0.52$M trainable parameters, the adaptation branch reduces ATE to $2.60$ m at $20$ epochs and $1.49$ m at $60$ epochs, consistently outperforming TartanIMU+LoRA with fewer updated parameters ($0.52$M vs. $0.74$M). 
The trajectory evolution in Figure~\ref{fig:finetune}(a) further confirms the efficiency of our adaptation mechanism.

Figure~\ref{fig:finetune}(b) evaluates source-domain retention after target-domain adaptation. 
The adapted model still maintains accurate trajectories on the original carrier domains, suggesting that freezing the base MosaicIMU model and learning only a residual branch mitigates catastrophic forgetting. 
We also conduct a router-adapter ablation, with results reported in Appendix~\ref{app:ablation_finetune}.

\begin{table*}[t!]
\centering
\caption{\textbf{Fine-tuning evaluation on the TLIO dataset.} ATE and RTE-10s are reported at different training epochs. Unit: m.}
\label{tab:finetune}
\renewcommand{\arraystretch}{1.3}
\resizebox{\linewidth}{!}{
\begin{tabular}{c|c|cc|cc|cc|cc}
\hline
\multirow{2}{*}{Method} & \multirow{2}{*}{\begin{tabular}{c}Trainable \\ Params.\end{tabular}} 
& \multicolumn{2}{c|}{0 Epoch} 
& \multicolumn{2}{c|}{20 Epoch} 
& \multicolumn{2}{c|}{40 Epoch} 
& \multicolumn{2}{c}{60 Epoch} \\
 & & ATE $\downarrow$ & RTE-10s $\downarrow$ 
   & ATE $\downarrow$ & RTE-10s $\downarrow$ 
   & ATE $\downarrow$ & RTE-10s $\downarrow$ 
   & ATE $\downarrow$ & RTE-10s $\downarrow$ \\
\hline
TartanIMU + LoRA               & 0.74M  & 54.27 & 16.78 & 18.82 & 3.70 & 12.09 & 2.53 & 5.97 & 2.26 \\
MosaicIMU + New Expert     & \textbf{0.52M}  & \textbf{9.85}  & \textbf{5.36}  & \textbf{2.60} & \textbf{1.32} & \textbf{2.22} & \textbf{0.88} & \textbf{1.49} & \textbf{0.72} \\
\hline
\end{tabular}}
\end{table*}

\subsection{Router-Guided Online Adaptation}
\label{sec:exp_online}
We finally evaluate router-guided online adaptation in a self-collected real-world vehicle experiment.
A four-wheeled vehicle is operated on campus roads, and the lightweight adaptation branch is updated online on a Jetson AGX Orin.
Following the online adaptation pipelines of~\citep{choi2026kiss} and~\citep{pan2024adaptive}, we use DLIO~\citep{chen2023direct} to provide velocity supervision, while the network input remains IMU-only.
During the warm-up, DLIO also supplies reference observations for filter updates to prevent state divergence from  the under-adapted model.
After convergence, DLIO is removed, and filter updates use the learned IMU-only velocity constraints. For the reported results below, each epoch checkpoint is evaluated by replaying the collected trajectory using IMU-only inference after online adaptation, with the DLIO estimate used as the reference trajectory for error computation.
Further implementation details are provided in Appendix~\ref{app:online_details}.

Figure~\ref{fig:online} shows that trajectory accuracy improves substantially during online adaptation. 
As the vehicle continues to travel, incoming data are used to progressively update the lightweight adaptation branch. 
The ATE decreases to $10.29$ m after $10$ epochs and further to $7.60$ m after $40$ epochs, while the RTE-10s drops from $10.52$ m to $2.12$ m. 
On a held-out test sequence from the same target domain, the adapted model achieves $9.73$ m ATE and $1.11$ m RTE-10s. 
This suggests that the selected samples capture transferable distribution shifts rather than merely overfitting to one trajectory.

\begin{figure*}[t]
    \centering
    \includegraphics[width=\textwidth]{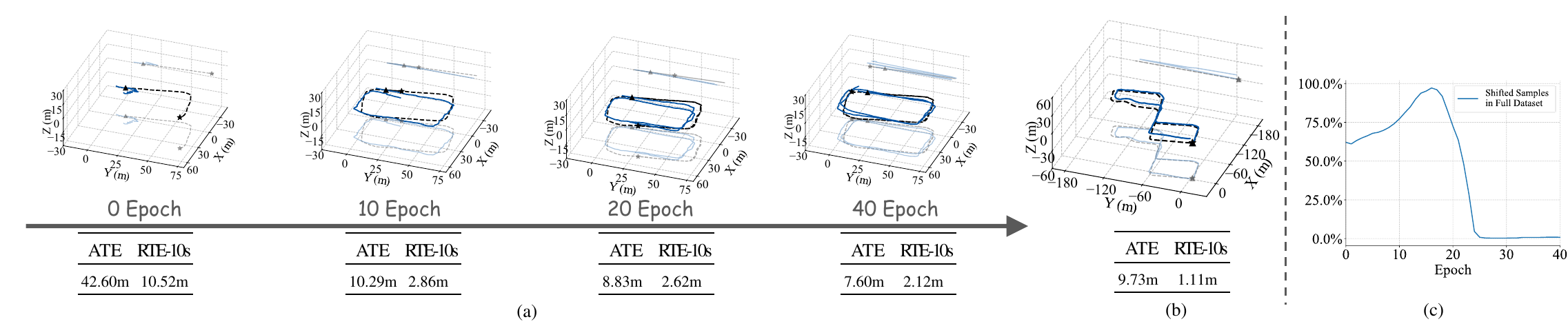}  
    \caption{\textbf{Router-guided online adaptation.} (a) The online fine-tuning process. (b) Evaluation on a held-out test trajectory not used for online fine-tuning. (c) The proportion of selected high-mismatch samples (shifted samples, weight on vehicle less than 0.8) decreases as the model adapts.}
    \label{fig:online}
\end{figure*}

The sample-selection trajectory further illustrates this adaptation behavior. 
The ratio of shifted samples starts at about $60\%$, increases briefly, and then drops to $0\%$ around the $25$th epoch. 
This indicates that as the model aligns with the target domain, incoming samples are no longer judged as highly mismatched. 
Continuing training until $40$ epochs further stabilizes the model and improves performance. 
Here, the router serves not only as an expert-assignment module, but also as a compact signal for selecting informative online samples under edge-computing constraints.

\section{Limitations}
\label{sec:limitations}
Despite the improved cross-carrier performance, MosaicIMU remains limited by the current data scale and carrier types, and thus is not expected to handle all possible IMU motion patterns. Larger domain shifts, including new IMU mountings within these categories, can degrade zero-shot performance, which also motivate the lightweight adaptation in Extensions I/II. Meanwhile, broader zero-shot transfer across different hardware or to unseen carrier types has not been systematically evaluated. The filter update also assumes reasonably stable IMU propagation and attitude estimation, so severe bias changes, unusual sensor mounting, and long-term deployment may lead to potential failure cases. In addition, the current online adaptation still requires reference supervision during the early adaptation stage. Future work will broaden the pretraining data, strengthen router uncertainty estimation, and explore more robust self-supervised online adaptation.

\section{Conclusion}
\label{sec:conclusion}
We present MosaicIMU, a carrier-conditioned MoE framework for generalizable inertial odometry across heterogeneous carriers. 
Instead of assigning each carrier to a fixed output head, MosaicIMU performs sample-adaptive feature-level expert composition and predicts local velocity constraints for a history-aware EKF. 
Experiments on four carrier categories show that our method achieves the best average generalization performance, reducing average ATE and RTE-10s by 40\% and 34\%, respectively, compared with the strongest learning-based baseline.
For unseen-domain deployment, the lightweight expert residual adaptation branch further improves the unseen TLIO dataset ATE to 1.49 m, while router-guided online adaptation reduces ATE from 42.60 m to 7.60 m using a compact buffer of selected windows. 
These results suggest that MosaicIMU offers a practical and scalable step toward general-purpose neural inertial odometry across heterogeneous carriers.

%===============================================================================

\clearpage
% The acknowledgments are automatically included only in the final and preprint versions of the paper.
\acknowledgments{This work is supported in part by the opening project from the State Key Laboratory of Autonomous Intelligent Unmanned Systems (No.  ZZKF2025ZD-2-2), in part by Tsinghua-Toyota Joint Research Fund, in part by National Natural Science Foundation of China (Grant No. 62633013). }

%===============================================================================

\bibliography{example}  % .bib

\clearpage
\appendix
\section*{Appendix}
\section{Method}

\subsection{Details of Training}
\label{app:training}
All models are implemented in PyTorch and trained on a single NVIDIA RTX 4090 GPU. 
The IMU data are resampled to 200 Hz and divided into sliding windows of $T=500$ frames with a stride of 250 frames. 
Each input window contains 6-D inertial measurements, including three-axis acceleration and angular velocity, and the ground-truth velocity is represented in the local gravity-aligned frame.

For the base model, MosaicIMU is trained on four carrier categories (vehicle, pedestrian, quadruped robot, and drone). 
We train the network for 100 epochs with a batch size of 128 using the Adam optimizer with a weight decay of $1\times10^{-4}$. 
The initial learning rate is set to $1\times10^{-3}$ and decayed to $1\times10^{-5}$ via cosine annealing.

\subsection{Details of History-Aware EKF}
\label{app:ekf}

This section provides the detailed formulation of the history-aware EKF used in Section~\ref{sec:ekf}. 
Our estimator builds on the error-state inertial filtering framework with cloned historical poses, 
which is a standard mechanism for processing delayed relative-state constraints in stochastic cloning 
and Multi-State Constraint Kalman Filter (MSCKF)-style estimators~\cite{roumeliotis2002stochastic,mourikis2007multi}.
Learned inertial odometry systems such as TLIO~\cite{liu2020tlio} and IMO~\cite{cioffi2023learned} also exploit this principle by using network-predicted relative displacement as a filter update. 
Nevertheless, this work does not regress a single displacement constraint. Instead, the proposed MosaicIMU predicts dense velocity and uncertainty sequences in the local gravity-aligned frame with its carrier-conditioned expert architecture.
We design a history-aware velocity-consistency interface that converts dense local predictions into a segment-level average-velocity observation between two cloned poses. 
This interface preserves the temporal richness of dense velocity regression, suppresses frame-wise prediction noise, and makes the neural output compatible with history-aware state estimation.

\paragraph{State and perturbation convention.}
We denote the current state as
\begin{equation}
    \mathbf{x}
    =
    \left(
    \mathbf{R}_{B}^{W},
    \mathbf{v}^{W},
    \mathbf{p}^{W},
    \mathbf{b}_{g},
    \mathbf{b}_{a}
    \right),
\end{equation}
where $\mathbf{R}_{B}^{W}\in SO(3)$ maps vectors from the body frame $B$ to the world frame $W$, 
$\mathbf{v}^{W}$ and $\mathbf{p}^{W}$ are the velocity and position in $W$, and 
$\mathbf{b}_{g},\mathbf{b}_{a}$ are the gyroscope and accelerometer biases. 
To support segment-level neural constraints, the filter maintains a bounded set of cloned historical poses,
\begin{equation}
    \mathcal{C}
    =
    \left\{
    \left(\mathbf{R}_{B_n}^{W}, \mathbf{p}_{n}^{W}\right)
    \right\}_{n=1}^{N}.
\end{equation}
The augmented state is
\begin{equation}
    \mathcal{X}
    =
    \left[
    \mathcal{C},
    \mathbf{R}_{B}^{W},
    \mathbf{v}^{W},
    \mathbf{p}^{W},
    \mathbf{b}_{g},
    \mathbf{b}_{a}
    \right].
\end{equation}
We use a left-multiplicative perturbation for rotations:
\begin{equation}
    \mathbf{R}
    \leftarrow
    \operatorname{Exp}(\delta\boldsymbol{\theta})\mathbf{R},
    \qquad
    \mathbf{p}
    \leftarrow
    \mathbf{p}+\delta\mathbf{p}.
\end{equation}
Accordingly, the augmented error state is
\begin{equation}
    \delta\mathcal{X}
    =
    \left[
    \delta\boldsymbol{\theta}_{1}^{\top},
    \delta\mathbf{p}_{1}^{\top},
    \cdots,
    \delta\boldsymbol{\theta}_{N}^{\top},
    \delta\mathbf{p}_{N}^{\top},
    \delta\boldsymbol{\theta}^{\top},
    \delta\mathbf{v}^{\top},
    \delta\mathbf{p}^{\top},
    \delta\mathbf{b}_{g}^{\top},
    \delta\mathbf{b}_{a}^{\top}
    \right]^{\top},
\end{equation}
and $\boldsymbol{\Sigma}\in\mathbb{R}^{(6N+15)\times(6N+15)}$ denotes the corresponding covariance. Here, $N$ is the number of currently stored clones.

\paragraph{IMU propagation.}
Let $\tilde{\boldsymbol{\omega}}_k$ and $\tilde{\mathbf{a}}_k$ be the gyroscope and accelerometer measurements at time $t_k$, and let $\Delta t=t_{k+1}-t_k$. 
After bias compensation,
\begin{equation}
    \boldsymbol{\omega}_k
    =
    \tilde{\boldsymbol{\omega}}_k-\mathbf{b}_{g,k},
    \qquad
    \mathbf{a}_k
    =
    \tilde{\mathbf{a}}_k-\mathbf{b}_{a,k}.
\end{equation}
The nominal state is propagated as
\begin{equation}
\begin{aligned}
    \Delta\boldsymbol{\theta}_k
    &=
    \boldsymbol{\omega}_k\Delta t,
    \\
    \mathbf{R}_{k+1}
    &=
    \mathbf{R}_{k}\operatorname{Exp}(\Delta\boldsymbol{\theta}_k),
    \\
    \mathbf{v}_{k+1}
    &=
    \mathbf{v}_{k}
    +
    \mathbf{R}_{k}\mathbf{a}_k\Delta t
    +
    \mathbf{g}\Delta t,
    \\
    \mathbf{p}_{k+1}
    &=
    \mathbf{p}_{k}
    +
    \mathbf{v}_{k}\Delta t
    +
    \frac{1}{2}\mathbf{R}_{k}\mathbf{a}_k\Delta t^2
    +
    \frac{1}{2}\mathbf{g}\Delta t^2,
    \\
    \mathbf{b}_{g,k+1}
    &=
    \mathbf{b}_{g,k},
    \qquad
    \mathbf{b}_{a,k+1}
    =
    \mathbf{b}_{a,k},
\end{aligned}
\end{equation}
where $\mathbf{g}=[0,0,-g]^{\top}$. 
The biases are kept constant in nominal propagation and modeled as random walks in covariance propagation.

For compactness, define
\begin{equation}
    \Delta\mathbf{v}_k
    =
    \mathbf{R}_{k}\mathbf{a}_k\Delta t,
    \qquad
    \Delta\mathbf{p}_k
    =
    \frac{1}{2}\mathbf{R}_{k}\mathbf{a}_k\Delta t^2.
\end{equation}
The current-state error transition matrix $\mathbf{A}_k\in\mathbb{R}^{15\times 15}$ is initialized to the identity matrix, with non-zero off-diagonal blocks
\begin{equation}
\begin{aligned}
    \mathbf{A}_{\mathbf{v}\boldsymbol{\theta}}
    &=
    -[\Delta\mathbf{v}_k]_{\times},
    &
    \mathbf{A}_{\mathbf{p}\boldsymbol{\theta}}
    &=
    -[\Delta\mathbf{p}_k]_{\times},
    \\
    \mathbf{A}_{\mathbf{p}\mathbf{v}}
    &=
    \mathbf{I}\Delta t,
    &
    \mathbf{A}_{\boldsymbol{\theta}\mathbf{b}_g}
    &=
    -\mathbf{R}_{k+1}\mathbf{J}_{r}(\Delta\boldsymbol{\theta}_k)\Delta t,
    \\
    \mathbf{A}_{\mathbf{v}\mathbf{b}_a}
    &=
    -\mathbf{R}_{k}\Delta t,
    &
    \mathbf{A}_{\mathbf{p}\mathbf{b}_a}
    &=
    -\frac{1}{2}\mathbf{R}_{k}\Delta t^2,
\end{aligned}
\end{equation}
where $[\cdot]_{\times}$ is the skew-symmetric matrix and $\mathbf{J}_{r}(\cdot)$ is the right Jacobian of $SO(3)$.

The IMU white-noise covariance and bias random-walk covariance are
\begin{equation}
    \mathbf{W}
    =
    \operatorname{diag}
    \left(
    \sigma_g^2\mathbf{I}_3,
    \sigma_a^2\mathbf{I}_3
    \right),
    \qquad
    \mathbf{Q}
    =
    \operatorname{diag}
    \left(
    \sigma_{b_g}^2\mathbf{I}_3,
    \sigma_{b_a}^2\mathbf{I}_3
    \right).
\end{equation}
The IMU noise Jacobian for the current state is
\begin{equation}
    \mathbf{B}_k
    =
    \begin{bmatrix}
    \mathbf{R}_{k+1}\mathbf{J}_{r}(\Delta\boldsymbol{\theta}_k)\Delta t & \mathbf{0} \\
    \mathbf{0} & \mathbf{R}_{k}\Delta t \\
    \mathbf{0} & \frac{1}{2}\mathbf{R}_{k}\Delta t^2 \\
    \mathbf{0} & \mathbf{0} \\
    \mathbf{0} & \mathbf{0}
    \end{bmatrix}.
\end{equation}

\paragraph{Pose cloning.}
Every $d$ IMU samples, the propagated current pose $(\mathbf{R}^{W}_{B_c}, \mathbf{p}^{W}_{c})$ is copied into the clone set.
Each neural observation spans $m$ clone intervals. Once the input buffer is full and $m+1$ clones are available, each newly added clone $j$ is paired with clone $i=j-m$ for one update, after which the oldest clone is removed.
Thus, updates occur at every clone insertion after warm-up, with $m$ clones retained between updates.

Since the cloned pose is generated from the current inertial propagation, its uncertainty and cross-correlations with the current state must be inherited from the propagation process. 
Let $\mathbf{A}_{c}\in\mathbb{R}^{15\times15}$ denote the error-state transition from the previous IMU state to the clone time. 
This transition maps the previous current-state error to the propagated 15-D current-state error. 
However, each clone only stores attitude and position, rather than the full inertial state. 
Therefore, the clone-augmentation Jacobian is obtained by selecting the attitude and position rows of $\mathbf{A}_{c}$:
\begin{equation}
    \mathbf{J}_{c}
    =
    \begin{bmatrix}
    \mathbf{A}_{c,\boldsymbol{\theta}} \\
    \mathbf{A}_{c,\mathbf{p}}
    \end{bmatrix}
    \in
    \mathbb{R}^{6\times 15}.
\end{equation}
With $N$ existing clones, the augmented transition matrix is
\begin{equation}
    \mathbf{A}^{aug}_{k}
    =
    \begin{bmatrix}
    \mathbf{I}_{6N} & \mathbf{0} \\
    \mathbf{0} & \mathbf{J}_{c} \\
    \mathbf{0} & \mathbf{A}_{k}
    \end{bmatrix},
\end{equation}
and the corresponding augmented IMU noise Jacobian is
\begin{equation}
    \mathbf{B}^{aug}_{k}
    =
    \begin{bmatrix}
    \mathbf{0}_{6N\times 6} \\
    \mathbf{B}_{c} \\
    \mathbf{B}_{k}
    \end{bmatrix},
\end{equation}
where $\mathbf{B}_{c}$ selects the attitude and position components of the propagation noise Jacobian. 
If no clone is inserted, $\mathbf{A}^{aug}_{k}$ reduces to an identity mapping for old clones and $\mathbf{A}_{k}$ for the current state.

The covariance is propagated by
\begin{equation}
    \boldsymbol{\Sigma}_{k+1}
    =
    \mathbf{A}^{aug}_{k}
    \boldsymbol{\Sigma}_{k}
    \mathbf{A}^{aug\top}_{k}
    +
    \mathbf{B}^{aug}_{k}
    \mathbf{W}
    \mathbf{B}^{aug\top}_{k}
    +
    \mathbf{G}_{b}
    \mathbf{Q}
    \mathbf{G}_{b}^{\top}
    \Delta t,
\end{equation}
where $\mathbf{G}_{b}$ selects the bias components of the current state. 
The propagated covariance preserves cross-correlations between historical poses and the current state.

\paragraph{Filter-aware aggregation of dense neural velocities.}
For an IMU window spanning the $m$ intervals between clones $i=j-m$ and $j$, the prediction length is chosen as $T=md$ to match the observation interval. MosaicIMU outputs dense local velocities and the corresponding variances,
\begin{equation}
    \left\{
    \left(
    \hat{\mathbf{v}}_{t}^{L_i},
    \hat{\mathbf{s}}_{t}^{L_i}
    \right)
    \right\}_{t=1}^{T},
\end{equation}
where $L_i$ denotes the local gravity-aligned frame associated with the start clone $i$, as defined in Section~\ref{sec:method}. 
Its vertical axis is aligned with gravity, while its yaw is anchored by the initial clone. The formulation below assumes that the neural velocities and the cloned-pose prediction use this same local-frame anchor.
Let $\psi_i$ be the yaw angle extracted from $\mathbf{R}_{B_i}^{W}$, and define
\begin{equation}
    \mathbf{R}_{z,i}
    =
    \begin{bmatrix}
    \cos\psi_i & -\sin\psi_i & 0 \\
    \sin\psi_i &  \cos\psi_i & 0 \\
    0          & 0           & 1
    \end{bmatrix},
    \qquad
    \mathbf{R}_{W}^{L_i}
    =
    \mathbf{R}_{z,i}^{\top}.
\end{equation}

We would like to remark that such a local gravity-aligned frame has two advantages. 
First, the measurement in the local gravity-aligned frame avoids adding artificial global-yaw information to the EKF update. 
In IMU-only estimation, global yaw cannot be directly observed. 
If the velocity is expressed in the world frame, its direction may be incorrectly treated as a global heading observation by the filter. 
Second, rotating IMU samples and velocity targets into a gravity-aligned frame provides the network with an implicit gravity reference. 
This allows MosaicIMU to learn carrier-dependent motion priors in a physically consistent coordinate system, while avoiding the instability of instantaneous body-frame targets.

Rather than treating each instantaneous velocity as an independent measurement, we introduce a segment aggregation operator
\begin{equation}
    \mathcal{M}_{ij}
    \left(
    \left\{
    \hat{\mathbf{v}}_{t}^{L_i}
    \right\}_{t\in\mathcal{I}_{ij}}
    \right)
    =
    \sum_{t\in\mathcal{I}_{ij}}
    \alpha_t
    \hat{\mathbf{v}}_{t}^{L_i},
    \qquad
    \sum_{t\in\mathcal{I}_{ij}}\alpha_t=1,
\end{equation}
where $\mathcal{I}_{ij}$ contains all $T$ prediction frames in the window associated with clones $i$ and $j$. For uniformly sampled IMU predictions, we set $\alpha_t=1/T$, so that each complete prediction window contributes one average-velocity observation.
The resulting neural observation is
\begin{equation}
    \mathbf{z}_{ij}
    =
    \mathcal{M}_{ij}
    \left(
    \left\{
    \hat{\mathbf{v}}_{t}^{L_i}
    \right\}_{t\in\mathcal{I}_{ij}}
    \right).
\end{equation}
Neglecting cross-covariances between frame-wise prediction errors, we approximate the observation covariance by aggregating and scaling the predicted variances over the same window:
\begin{equation}
    \mathbf{N}_{ij}
    =
    \sum_{t\in\mathcal{I}_{ij}}
    \alpha_t^2
    \operatorname{diag}
    \left(
    \hat{\mathbf{s}}_{t}^{L_i}
    \right).
\end{equation}

\paragraph{Velocity-consistency measurement model.}
The filter prediction corresponding to $\mathbf{z}_{ij}$ is derived from the average motion of the two cloned poses. 
Define the world-frame segment velocity induced by the cloned poses as
\begin{equation}
    \bar{\mathbf{v}}_{ij}^{W}
    =
    \frac{
    \mathbf{p}_{j}^{W}
    -
    \mathbf{p}_{i}^{W}
    }{\Delta t_{ij}},
    \qquad
    \Delta t_{ij}=t_j-t_i.
\end{equation}
Note that this quantity is not obtained by directly integrating the network prediction, but is a state-dependent velocity implied by the historical trajectory maintained by the filter. 
Projecting it into the local gravity-aligned frame gives
\begin{equation}
    h_{ij}(\mathcal{X})
    =
    \mathbf{R}_{W}^{L_i}
    \bar{\mathbf{v}}_{ij}^{W}
    =
    \mathbf{R}_{z,i}^{\top}
    \frac{
    \mathbf{p}_{j}^{W}
    -
    \mathbf{p}_{i}^{W}
    }{\Delta t_{ij}}.
\end{equation}
The residual is therefore
\begin{equation}
    \mathbf{r}_{ij}
    =
    \mathbf{z}_{ij}
    -
    h_{ij}(\mathcal{X}).
\end{equation}

This residual defines a velocity-consistency constraint between the dense neural velocity field and the historical trajectory estimated by the EKF.

\paragraph{Yaw consistency.}
Although $h_{ij}$ depends on the initial clone yaw through $R_{z,i}$, this yaw only defines the local gravity-aligned frame of the network observation and does not provide an absolute-yaw measurement. Therefore, for any global yaw rotation $R_\psi$, applying
$p_i^W \leftarrow R_\psi p_i^W$, $p_j^W \leftarrow R_\psi p_j^W$, and $R_{z,i} \leftarrow R_\psi R_{z,i}$ leaves the measurement unchanged:
\[
(R_\psi R_{z,i})^\top
\frac{R_\psi(p_j^W-p_i^W)}{\Delta t_{ij}}
=
R_{z,i}^\top
\frac{p_j^W-p_i^W}{\Delta t_{ij}}.
\]
This means that the residual is independent of the arbitrary global-yaw reference when the neural observation is expressed in the same local frame. The measurement model therefore enforces consistency between the cloned-pose displacement and the local average velocity without introducing an explicit absolute-heading observation.

\paragraph{Measurement Jacobian.}
The measurement function depends on the initial clone yaw only through the local-frame anchor, and on the positions of the start and end clones.
For compactness, denote
\begin{equation}
    \mathbf{q}_{ij}
    =
    \bar{\mathbf{v}}_{ij}^{W}
    =
    \frac{\mathbf{p}_{j}^{W}-\mathbf{p}_{i}^{W}}{\Delta t_{ij}},
    \qquad
    \hat{\mathbf{z}}_{ij}
    =
    \mathbf{R}_{z,i}^{\top}\mathbf{q}_{ij}.
\end{equation}
The position Jacobian blocks are
\begin{equation}
    \mathbf{H}_{\mathbf{p}_i}
    =
    -\frac{\mathbf{R}_{z,i}^{\top}}{\Delta t_{ij}},
    \qquad
    \mathbf{H}_{\mathbf{p}_j}
    =
    \frac{\mathbf{R}_{z,i}^{\top}}{\Delta t_{ij}}.
\end{equation}

We derive the orientation block through the yaw anchoring of the local frame. 
Let $\mathbf{e}_z=[0,0,1]^{\top}$. 
A small perturbation of the initial clone yaw produces
\begin{equation}
    \delta
    \left(
    \mathbf{R}_{z,i}^{\top}\mathbf{q}_{ij}
    \right)
    =
    \left[
    \mathbf{R}_{z,i}^{\top}\mathbf{q}_{ij}
    \right]_{\times}
    \mathbf{e}_z
    \delta\psi_i.
\end{equation}
The yaw perturbation is related to the left attitude perturbation by
\begin{equation}
    \delta\psi_i
    =
    \mathbf{j}_{\psi,i}
    \delta\boldsymbol{\theta}_i,
\end{equation}
where $\mathbf{j}_{\psi,i}\in\mathbb{R}^{1\times3}$ is the yaw-extraction Jacobian evaluated at $\mathbf{R}_{B_i}^{W}$. 
For a yaw-pitch-roll parameterization with pitch $\vartheta_i$ and yaw $\psi_i$, it can be obtained that
\begin{equation}
    \mathbf{j}_{\psi,i}
    =
    \begin{bmatrix}
    \cos\psi_i\tan\vartheta_i &
    \sin\psi_i\tan\vartheta_i &
    1
    \end{bmatrix}.
\end{equation}
Thus, the initial orientation block is
\begin{equation}
    \mathbf{H}_{\boldsymbol{\theta}_i}
    =
    \left[
    \hat{\mathbf{z}}_{ij}
    \right]_{\times}
    \mathbf{e}_z
    \mathbf{j}_{\psi,i}.
\end{equation}
All other blocks are zero:
\begin{equation}
    \mathbf{H}_{ij}
    =
    \begin{bmatrix}
    \cdots &
    \mathbf{H}_{\boldsymbol{\theta}_i} &
    \mathbf{H}_{\mathbf{p}_i} &
    \cdots &
    \mathbf{0}_{3\times 3} &
    \mathbf{H}_{\mathbf{p}_j} &
    \cdots
    \end{bmatrix}.
\end{equation}
This sparse Jacobian makes the update efficient while preserving the historical correlations required by the segment-level observation.

\paragraph{Kalman update and marginalization.}
Given $\mathbf{r}_{ij}$ and $\mathbf{H}_{ij}$, the innovation covariance and Kalman gain are
\begin{equation}
    \mathbf{S}_{ij}
    =
    \mathbf{H}_{ij}
    \boldsymbol{\Sigma}
    \mathbf{H}_{ij}^{\top}
    +
    \mathbf{N}_{ij},
    \qquad
    \mathbf{K}_{ij}
    =
    \boldsymbol{\Sigma}
    \mathbf{H}_{ij}^{\top}
    \mathbf{S}_{ij}^{-1}.
\end{equation}
The error-state correction is
\begin{equation}
    \delta\mathcal{X}
    =
    \mathbf{K}_{ij}
    \mathbf{r}_{ij}.
\end{equation}
The correction is injected into both cloned poses and the current state:
\begin{equation}
\begin{aligned}
    \mathbf{R}_{B_n}^{W}
    &\leftarrow
    \operatorname{Exp}(\delta\boldsymbol{\theta}_{n})\mathbf{R}_{B_n}^{W},
    &
    \mathbf{p}_{n}^{W}
    &\leftarrow
    \mathbf{p}_{n}^{W}+\delta\mathbf{p}_{n},
    \\
    \mathbf{R}_{B}^{W}
    &\leftarrow
    \operatorname{Exp}(\delta\boldsymbol{\theta})\mathbf{R}_{B}^{W},
    &
    \mathbf{v}^{W}
    &\leftarrow
    \mathbf{v}^{W}+\delta\mathbf{v},
    \\
    \mathbf{p}^{W}
    &\leftarrow
    \mathbf{p}^{W}+\delta\mathbf{p},
    &
    \mathbf{b}_{g}
    &\leftarrow
    \mathbf{b}_{g}+\delta\mathbf{b}_{g},
    \quad
    \mathbf{b}_{a}
    \leftarrow
    \mathbf{b}_{a}+\delta\mathbf{b}_{a}.
\end{aligned}
\end{equation}
The covariance is updated as
\begin{equation}
    \boldsymbol{\Sigma}
    \leftarrow
    \boldsymbol{\Sigma}
    -
    \mathbf{K}_{ij}\mathbf{H}_{ij}\boldsymbol{\Sigma}.
\end{equation}
The covariance is then symmetrized:
\begin{equation}
    \boldsymbol{\Sigma}
    \leftarrow
    \frac{1}{2}
    \left(
    \boldsymbol{\Sigma}
    +
    \boldsymbol{\Sigma}^{\top}
    \right).
\end{equation}
After each update, the oldest clone $i=j-m$ is marginalized by deleting its state entries and the corresponding covariance rows and columns, leaving the $m$ most recent clones for the next update.
Although the observation directly involves only the two endpoint clones, the Kalman correction also adjusts the intermediate clones and the current inertial state through their cross-covariances.
This keeps the filter size bounded while retaining the historical correlations needed for subsequent velocity-consistency updates.

\paragraph{Discussion.}
The proposed filter is designed to bridge dense velocity prediction and probabilistic inertial state estimation. 
Dense predictions preserve fine-grained temporal motion cues during learning, while the segment-level average-velocity observation provides a robust and estimator-compatible constraint during inference.  In addition, such a design is particularly suitable for heterogeneous carriers: instantaneous velocity estimates may be affected by carrier-specific vibration, contact, or aerodynamic disturbances, whereas their segment-level velocity consistency remains a stable cue for correcting inertial drift. 
As a result, the EKF acts as a history-aware integration layer that converts MosaicIMU's learned local motion priors into corrected inertial trajectories.

\section{Experimental Results}

\subsection{Multi-Carrier IMU Training Datasets}
\label{app:dataset_details}

To evaluate multi-carrier generalization in neural inertial odometry across heterogeneous carriers, we construct a multi-carrier IMU benchmark from several publicly available datasets. As summarized in Table~\ref{tab:datasets}, the training datasets cover four representative carrier types, including vehicles, quadruped robots, pedestrians, and drones. To mitigate bias toward dominant motion categories, we standardize the training distribution by re-weighting samples from different carriers. This ensures that each carrier group contributes approximately the same proportion of training samples.

To avoid overlap leakage, all data splits are performed before sliding-window extraction. For each dataset, complete trajectories are first divided into training and held-out test trajectories with a 9:1 ratio. The held-out test trajectories are never used during training, validation, early stopping, hyperparameter tuning, or normalization-statistic estimation. For each training trajectory, the first 5\% and the last 15\% of data are reserved as validation segments, while the middle 80\% is used for training. Sliding windows are generated separately after temporal segmentation. Windows crossing train/validation boundaries are discarded, and a one-window-length margin is removed around each split point to prevent raw IMU samples from being shared across training and validation windows. All raw IMU sequences are resampled to 200 Hz, and each valid segment is divided into sliding windows of 500 frames with a stride of 250 frames.

During training, we apply lightweight but physically consistent data augmentation. Specifically, zero-mean Gaussian white noise is added to the IMU measurements to improve robustness against sensor noise. In addition, we randomly rotate each IMU window around the gravity axis, together with its corresponding velocity labels, to augment heading variations while preserving the gravity direction and local inertial dynamics.

\begin{table}[h]
    \centering
    \caption{\textbf{Multi-carrier IMU datasets used for training and evaluation.}}
    \label{tab:datasets}
    \renewcommand{\arraystretch}{1.2}
    \resizebox{0.6\textwidth}{!}{
        \begin{tabular}{cccc} 
            \toprule
            \textbf{Carrier} & \textbf{Dataset} & \textbf{Duration} & \textbf{Environment} \\ 
            \midrule
            \multirow{2}{*}{Vehicle} 
            & i2Nav-Robot~\citep{tang2025i2nav} & 4h & Indoors, Outdoors \\
            & SubT-MRS~\citep{zhao2024subt} & 3h & In/Outdoors, Underground \\ 
            \hline

            \multirow{2}{*}{Quadruped} 
            & CEAR~\citep{zhu2024cear} & 4h & Indoors, Outdoors \\
            & Cerberus~\citep{yang2022cerberus} & 2h & Underground \\ 
            \hline

            \multirow{2}{*}{Pedestrian} 
            & IDOL~\citep{sun2021idol} & 5.4h & Indoors \\
            & RNIN-VIO~\citep{chen2021rnin} & 5h & Indoors, Outdoors \\ 
            \hline

            \multirow{2}{*}{Drone} 
            & BlackBird~\citep{antonini2018blackbird} & $<$1h & Indoors \\
            & MILUV~\citep{shalaby2025miluv} & 3.6h & Indoors \\ 
            \bottomrule
        \end{tabular}
    }
\end{table}

\subsection{Details of Compared Baselines}
\label{app:baseline}

To validate the generalization ability of MosaicIMU, we compare it with three representative open-source learning-based inertial odometry baselines: IMO~\citep{cioffi2023learned}, TLIO~\citep{liu2020tlio}, and TartanIMU~\citep{zhao2025tartan}. These methods cover carrier-specific neural inertial estimation, filter-based neural-inertial fusion, and a recent multi-carrier IMU pretrained model. For fair evaluation, IMO and TLIO are trained separately on the corresponding carrier data, while TartanIMU and MosaicIMU are trained on the same multi-carrier pretraining datasets. All four methods are evaluated on identical held-out sequences.
\begin{itemize}
    \item \textbf{IMO}~\citep{cioffi2023learned} is designed for autonomous drone racing. It combines an IMU-propagated EKF with a temporal convolutional network that predicts relative 3-DoF displacement from collective thrust and gyroscope measurements. The predicted displacement is then used as an EKF update. We include it as a representative drone-specific learned inertial odometry baseline. However, because IMO does not enforce rigid kinematic constraints, it can effectively adapt to various agents by simply training dedicated models on the respective carrier's data.

    \item \textbf{TLIO}~\citep{liu2020tlio} predicts local inertial displacement and uncertainty from short IMU windows, and incorporates the learned motion constraint into an error-state Kalman filter. Unlike cross-carrier models, TLIO learns a single motion prior and is not explicitly designed to compose heterogeneous carrier dynamics.

    \item \textbf{TartanIMU}~\citep{zhao2025tartan} learns IMU representations from heterogeneous carriers and uses carrier-specific output heads for different platforms. It is the closest baseline to our model, but its fixed-head design requires explicit carrier assignment and does not support sample-adaptive expert composition. Since TartanIMU predicts velocities in its local body frame, we fuse its outputs using a standard velocity-observation EKF rather than the proposed history-aware EKF to obtain trajectories.
\end{itemize}

\paragraph{Excluded methods.}
We do not include some other inertial odometry methods whose assumptions are less compatible with our proposed model. For example, AI-IMU~\citep{brossard2020ai} introduces vehicle-specific motion assumptions such as non-holonomic constraints, which are effective for ground vehicles but difficult to generalize to pedestrians, quadruped robots, and drones. RoNIN~\citep{herath2020ronin} relies on transforming inertial predictions with an approximate rotation estimate, making it sensitive to the availability and quality of external attitude priors. Since our goal is to evaluate generalizable IMU-only velocity learning across heterogeneous carriers, we focus on the three open-source baselines above that can be adapted consistently to all evaluated platforms.

\subsection{Evaluation Metrics}
\label{app:metrics}

We compute all metrics using the estimated positions $\hat{\mathbf{p}}_t^W$ and ground-truth positions $\mathbf{p}_t^W$ in the world frame $W$.

\paragraph{Absolute Trajectory Error (ATE).}
ATE measures the root mean squared position error over the full trajectory:
\begin{equation}
    \mathrm{ATE}
    =
    \sqrt{\frac{1}{M}\sum_{t=1}^{M}
    \|\hat{\mathbf{p}}_t^W-\mathbf{p}_t^W\|_2^2},
\end{equation}
where $M$ is the number of evaluated timestamps.

\paragraph{Relative Trajectory Error over 10 seconds (RTE-10s).}
RTE-10s measures local drift by comparing estimated and ground-truth displacements over 10-second intervals:
\begin{equation}
    \mathrm{RTE\text{-}10s}
    =
    \sqrt{\frac{1}{M-\Delta}
    \sum_{t=1}^{M-\Delta}
    \left\|
    (\hat{\mathbf{p}}_{t+\Delta}^W-\hat{\mathbf{p}}_t^W)
    -
    (\mathbf{p}_{t+\Delta}^W-\mathbf{p}_t^W)
    \right\|_2^2},
\end{equation}
where $\Delta$ is the number of sampling intervals corresponding to 10 seconds.

\paragraph{Vertical trajectory error ($z$-Drift).}
$z$-Drift is ATE computed along the world-frame $z$-axis:
\begin{equation}
    z\mathrm{\text{-}Drift}
    =
    \sqrt{\frac{1}{M}\sum_{t=1}^{M}
    (\hat{p}_{t,z}^W-p_{t,z}^W)^2}.
\end{equation}
All three metrics are reported in meters.

\subsection{Ablation Studies of Pretrained MosaicIMU}
\label{app:ablation_pretrain}

\begin{figure*}[t] % 或者使用 [p] 如果它们占满了整整一页
    \centering
    
    % ==========================================
    % 1. 顶部的图片
    % ==========================================
    \includegraphics[width=0.9\linewidth]{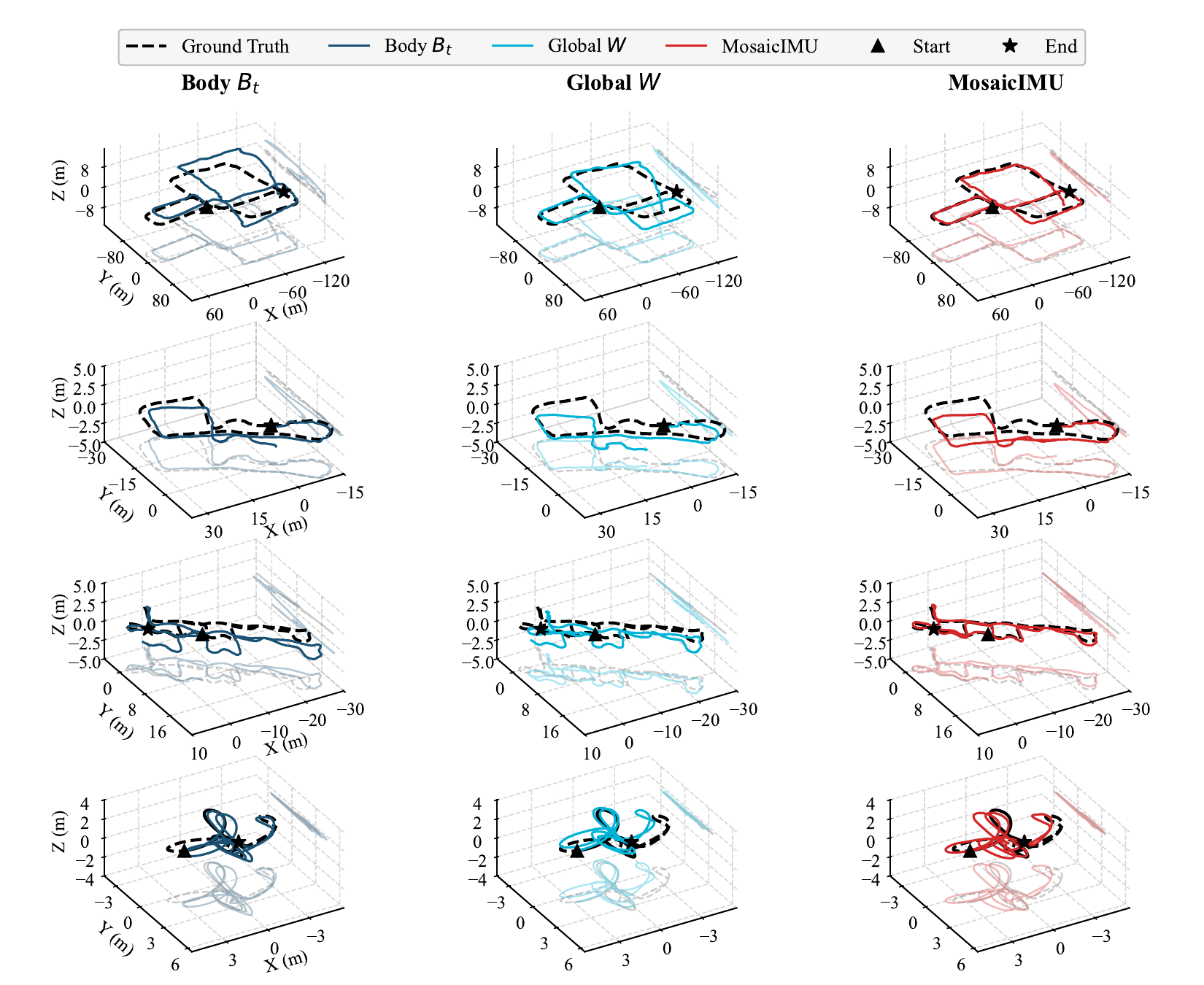}
    \caption{\textbf{Trajectory comparison under different velocity prediction frames.} 
From top to bottom, the four rows correspond to vehicle, quadruped robot, pedestrian, and drone sequences, respectively. }
    \label{fig:app_frame_ablation}

    % ==========================================
    % 2. 中间的大表格
    % ==========================================
\captionsetup{type=table} % 告诉 LaTeX 这里的 caption 属于 Table
    \caption{\textbf{Generalization evaluation under different velocity prediction frames.} 
Unit: m.}
    \label{tab:app_frame_ablation}
    \renewcommand{\arraystretch}{1.2}
    \setlength{\arrayrulewidth}{1pt}
    \resizebox{0.9\linewidth}{!}{%
        \begin{tabular}{c|ccc|ccc|ccc} 
        \hline
        \multirow{2}{*}{\textbf{Carrier}} 
        & \multicolumn{3}{c|}{\textbf{Body $B_t$}} 
        & \multicolumn{3}{c|}{\textbf{Global $W$}} 
        & \multicolumn{3}{c}{\textbf{MosaicIMU $L_0$}} \\
        & ATE $\downarrow$ & RTE-10s $\downarrow$ & $z$-Drift $\downarrow$ 
        & ATE $\downarrow$ & RTE-10s $\downarrow$ & $z$-Drift $\downarrow$ 
        & ATE $\downarrow$ & RTE-10s $\downarrow$ & $z$-Drift $\downarrow$ \\ 
        \hline

        Vehicle     
        & 14.64 & 3.26 & 3.68 
        & 26.76 & 3.95 & 1.91 
        & \cellcolor[rgb]{1,1,0.741}\textbf{10.56 }\textcolor[rgb]{0,0.502,0}{28\%$\downarrow$} 
        & \cellcolor[rgb]{1,1,0.741}\textbf{2.17 }\textcolor[rgb]{0,0.502,0}{33\%$\downarrow$} 
        & \cellcolor[rgb]{1,1,0.741}\textbf{1.50 }\textcolor[rgb]{0,0.502,0}{21\%$\downarrow$} \\

        Quadruped  
        & 2.44 & 0.69 & 1.38 
        & 2.55 & 0.93 & 1.56 
        & \cellcolor[rgb]{1,1,0.741}\textbf{2.01 }\textcolor[rgb]{0,0.502,0}{18\%$\downarrow$} 
        & \cellcolor[rgb]{1,1,0.741}\textbf{0.63 }\textcolor[rgb]{0,0.502,0}{9\%$\downarrow$} 
        & \cellcolor[rgb]{1,1,0.741}\textbf{0.86 }\textcolor[rgb]{0,0.502,0}{38\%$\downarrow$} \\

        Pedestrian   
        & 13.47 & 2.16 & 1.92 
        & 9.12 & 3.89 & 0.65 
        & \cellcolor[rgb]{1,1,0.741}\textbf{2.85 }\textcolor[rgb]{0,0.502,0}{69\%$\downarrow$} 
        & \cellcolor[rgb]{1,1,0.741}\textbf{1.28 }\textcolor[rgb]{0,0.502,0}{41\%$\downarrow$} 
        & \cellcolor[rgb]{1,1,0.741}\textbf{0.35 }\textcolor[rgb]{0,0.502,0}{46\%$\downarrow$} \\

        Drone       
        & 1.44 & 1.62 & 0.92 
        & 3.17 & 2.18 & 0.25 
        & \cellcolor[rgb]{1,1,0.741}\textbf{1.17 }\textcolor[rgb]{0,0.502,0}{19\%$\downarrow$} 
        & \cellcolor[rgb]{1,1,0.741}\textbf{1.14 }\textcolor[rgb]{0,0.502,0}{30\%$\downarrow$} 
        & \cellcolor[rgb]{1,1,0.741}\textbf{0.22 }\textcolor[rgb]{0,0.502,0}{12\%$\downarrow$} \\ 

        \hline

        Average     
        & 8.00 & 1.93 & 1.98 
        & 10.40 & 2.74 & 1.09 
        & \cellcolor[rgb]{1,1,0.741}\textbf{4.15 }\textcolor[rgb]{0,0.502,0}{48\%$\downarrow$} 
        & \cellcolor[rgb]{1,1,0.741}\textbf{1.31 }\textcolor[rgb]{0,0.502,0}{32\%$\downarrow$} 
        & \cellcolor[rgb]{1,1,0.741}\textbf{0.73 }\textcolor[rgb]{0,0.502,0}{33\%$\downarrow$} \\

        \hline
        \end{tabular}%
    }

\end{figure*}

\paragraph{Ablation on prediction in different frames.}
We further evaluate velocity prediction in different frames in Figure~\ref{fig:app_frame_ablation} and Table~\ref{tab:app_frame_ablation}. 
The global-frame variant predicts velocity in the world frame $W$, while the body-frame variant uses the instantaneous body frame $B_t$. 
The proposed algorithm predicts velocity in the local gravity-aligned frame $L_0$, which keeps a stable local reference while decoupling the observation from global yaw.

The $L_0$ representation achieves the best overall performance across all evaluated carriers and metrics. 
As reported in Table~\ref{tab:app_frame_ablation}, it reduces the average ATE from $8.00$ m to $4.15$ m compared with the $B_t$ and $W$ variants, and lowers the average RTE-10s from $1.93$ m to $1.31$ m. 
The improvement is particularly evident on vehicle and pedestrian sequences, where $L_0$ substantially reduces large trajectory deviations. 
Figure~\ref{fig:app_frame_ablation} further shows that the two alternative frames fail in different ways: global-frame prediction can couple velocity observations with attitude errors from the EKF, leading to accumulated trajectory drift, whereas body-frame prediction changes continuously with the rotation of the body, making dense velocity regression less stable. 
In contrast, the local gravity-aligned frame $L_0$ provides a fixed and gravity-consistent reference for each window, yielding more stable velocity learning and more reliable segment-level EKF updates without introducing spurious global-yaw constraints.

\begin{figure*}[p] % 或者使用 [p] 如果它们占满了整整一页
    \centering
    
    % ==========================================
    % 1. 顶部的图片
    % ==========================================
    \includegraphics[width=0.9\linewidth]{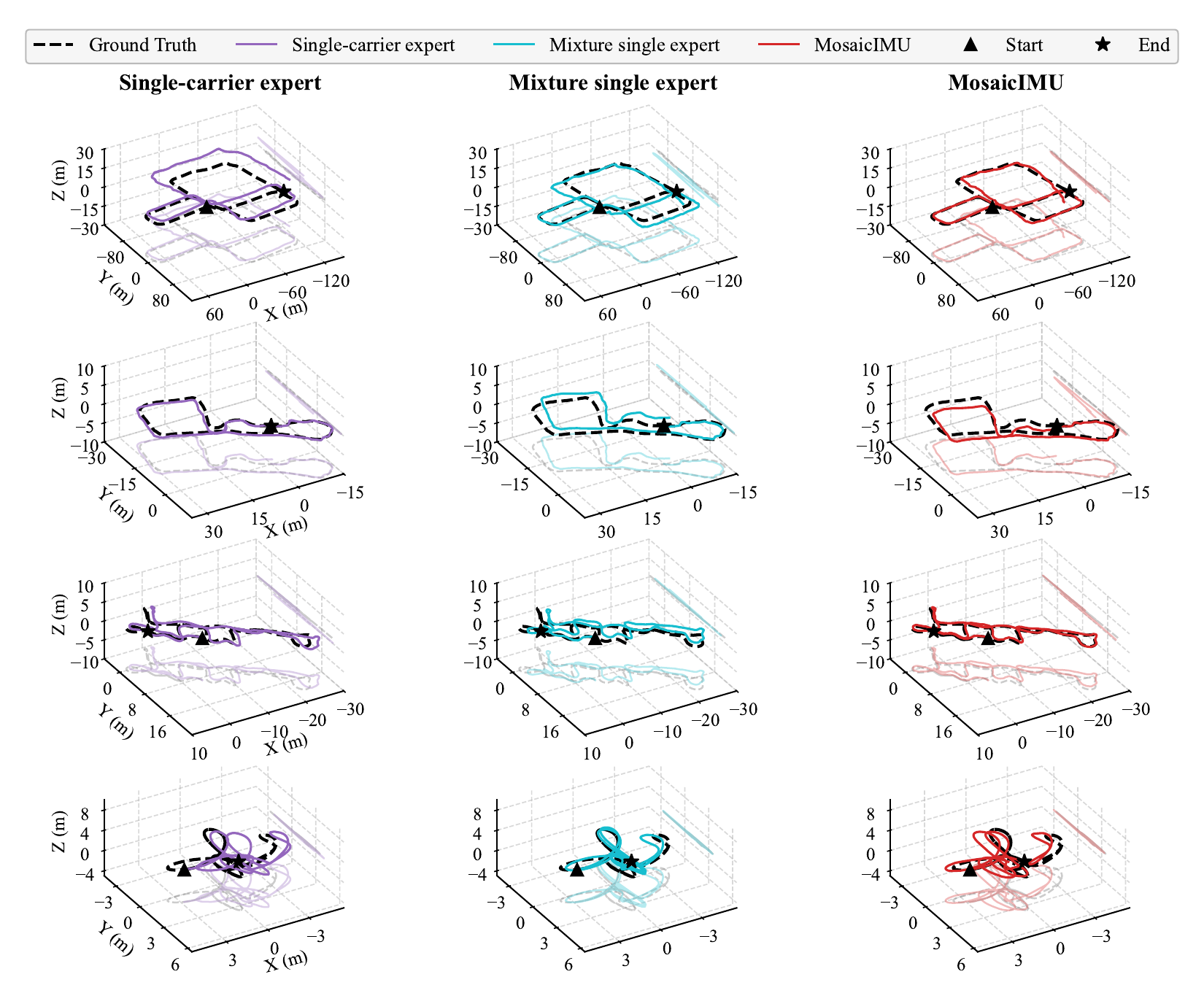}
    \caption{\textbf{Comparison of estimated trajectories from different expert structures.} From top to bottom, the four rows correspond to vehicle, quadruped robot, pedestrian, and drone sequences, respectively.}
    \label{fig:app_expert_ablation}
    
    \vspace{0.2cm} % 添加图片与表格之间的垂直间距
    
    % ==========================================
    % 2. 中间的大表格
    % ==========================================
    \begingroup % 开启分组，防止 captionsetup 污染后续图片的 caption
   \captionsetup{type=table} % 告诉 LaTeX 这里的 caption 属于 Table
\caption{\textbf{Generalization evaluation of different expert designs.} Unit: m.}
\label{tab:app_expert_ablation}
\renewcommand{\arraystretch}{1.2}
\setlength{\arrayrulewidth}{1pt}
\resizebox{0.65\linewidth}{!}{%
    \begin{tabular}{c|cc|cc|cc} 
    \hline
    \multirow{2}{*}{\textbf{Carrier}} 
    & \multicolumn{2}{c|}{\textbf{Single-carrier expert}} 
    & \multicolumn{2}{c|}{\textbf{Mixture single expert}} 
    & \multicolumn{2}{c}{\textbf{MosaicIMU}} \\
    & ATE $\downarrow$ & RTE-10s $\downarrow$ 
    & ATE $\downarrow$ & RTE-10s $\downarrow$ 
    & ATE $\downarrow$ & RTE-10s $\downarrow$ \\ 
    \hline

    Vehicle     
    & 14.12 & 3.09 
    & 14.14 & 3.21 
    & \cellcolor[rgb]{1,1,0.741}\textbf{10.56} \textcolor[rgb]{0,0.502,0}{25\%$\downarrow$} 
    & \cellcolor[rgb]{1,1,0.741}\textbf{2.17} \textcolor[rgb]{0,0.502,0}{30\%$\downarrow$} \\

    Quadruped  
    & 2.14 & 1.37 
    & 2.42 & 0.69 
    & \cellcolor[rgb]{1,1,0.741}\textbf{2.01} \textcolor[rgb]{0,0.502,0}{6\%$\downarrow$} 
    & \cellcolor[rgb]{1,1,0.741}\textbf{0.63} \textcolor[rgb]{0,0.502,0}{9\%$\downarrow$} \\

    Pedestrian   
    & 3.99 & 2.45 
    & 4.25 & 1.82 
    & \cellcolor[rgb]{1,1,0.741}\textbf{2.85} \textcolor[rgb]{0,0.502,0}{29\%$\downarrow$} 
    & \cellcolor[rgb]{1,1,0.741}\textbf{1.28} \textcolor[rgb]{0,0.502,0}{30\%$\downarrow$} \\

    Drone       
    & 1.61 & 1.45 
    & 1.31 & 1.22 
    & \cellcolor[rgb]{1,1,0.741}\textbf{1.17} \textcolor[rgb]{0,0.502,0}{11\%$\downarrow$} 
    & \cellcolor[rgb]{1,1,0.741}\textbf{1.14} \textcolor[rgb]{0,0.502,0}{7\%$\downarrow$} \\ 

    \hline

    Average     
    & 5.47 & 2.09 
    & 5.53 & 1.74 
    & \cellcolor[rgb]{1,1,0.741}\textbf{4.15} \textcolor[rgb]{0,0.502,0}{24\%$\downarrow$} 
    & \cellcolor[rgb]{1,1,0.741}\textbf{1.31} \textcolor[rgb]{0,0.502,0}{25\%$\downarrow$} \\

    \hline
    \end{tabular}%
}
    \endgroup % 结束分组
    \vspace{0.5cm}

    \centering
    
    % width=0.5\linewidth 表示图片宽度为当前文本行宽的 50%，你可以根据需要调整大小（如 0.6, 0.8 等）
    \includegraphics[width=0.4\linewidth]{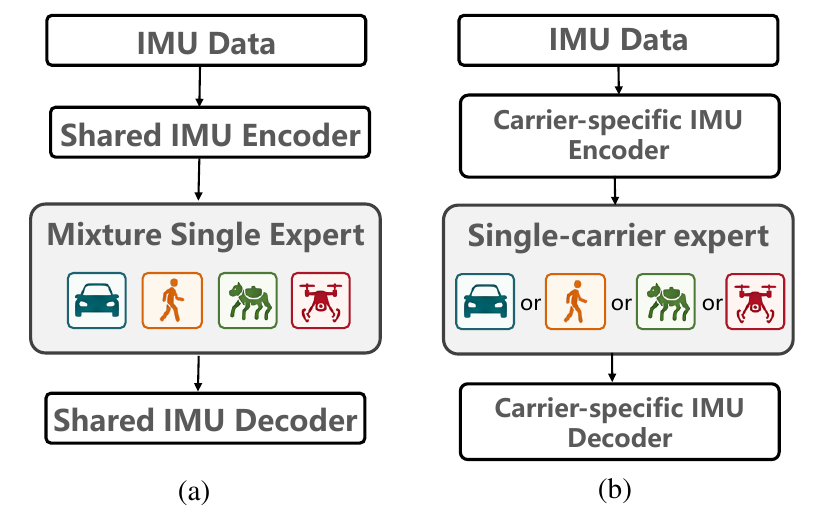}
    
    \caption{\textbf{Structural comparison of the baseline expert designs.} (a) Mixture single expert, and (b) Single-carrier expert. Note that both frameworks employ only a single expert network without a routing module.}
    \label{fig:app_expert_comparison}

\end{figure*}

\paragraph{Ablation on expert design.}
Table~\ref{tab:app_expert_ablation} and Figure~\ref{fig:app_expert_ablation} report per-carrier results for different expert designs. 
As shown in Figure \ref{fig:app_expert_comparison}, the single-carrier expert baseline trains four independent single-expert networks, each using data from only one carrier. 
The mixture single expert uses one expert trained on all carriers without any carrier-specific branch. 
MosaicIMU uses the router to adaptively combine expert features for each input window.

As shown in Table~\ref{tab:app_expert_ablation}, MosaicIMU achieves the best ATE and RTE-10s across all carriers. 
Compared with the single-carrier expert baseline, it reduces the average ATE from 5.47\,m to 4.15\,m and the average RTE-10s from 2.09\,m to 1.31\,m, suggesting that isolated carrier-specific training limits cross-carrier knowledge transfer. 
The mixture single expert has a higher average ATE of 5.53\,m, showing that a single shared expert is insufficient to capture heterogeneous inertial dynamics. 
By contrast, MosaicIMU combines shared representation learning with carrier-aware expert specialization. 
These results verify the importance of sample-adaptive feature-level composition for multi-carrier inertial learning.

\paragraph{Ablation on soft routing weights.}
Table~\ref{tab:app_routing_ablation} evaluates whether MosaicIMU benefits from soft expert composition rather than hard carrier-specific routing. 
The one-hot variant activates only the expert associated with the carrier label, which preserves carrier specialization but prevents information sharing across experts. 
In contrast, MosaicIMU uses the router weights to softly compose expert features for each input window.

As shown in Table~\ref{tab:app_routing_ablation}, soft routing improves the average ATE from 4.71\,m to 4.15\,m and the average RTE-10s from 1.34\,m to 1.31\,m. 
The improvement is consistent in ATE across all carriers, indicating that the routed experts are not merely independent carrier-specific modules. 
Instead, the router keeps the dominant carrier expert while allowing useful complementary features from other experts to participate in the prediction. 
The overall results show that soft feature-level composition provides a more flexible mechanism for heterogeneous inertial dynamics than a rigid one-hot expert assignment.

\begin{table*}[h]
\centering
\caption{\textbf{Ablation on soft routing weights.}
We compare MosaicIMU with a one-hot routing variant that activates only the carrier-specific expert. Unit: m.}
\label{tab:app_routing_ablation}
\renewcommand{\arraystretch}{1.2}
\setlength{\arrayrulewidth}{1pt}
\resizebox{0.5\linewidth}{!}{%
    \begin{tabular}{c|cc|cc} 
    \hline
    \multirow{2}{*}{\textbf{Carrier}} 
    & \multicolumn{2}{c|}{\textbf{MosaicIMU} (one-hot)} 
    & \multicolumn{2}{c}{\textbf{MosaicIMU}} \\
    & ATE $\downarrow$ & RTE-10s $\downarrow$ 
    & ATE $\downarrow$ & RTE-10s $\downarrow$ \\ 
    \hline

    Vehicle     
    & 12.40 & 2.31 
    & \cellcolor[rgb]{1,1,0.741}\textbf{10.56} \textcolor[rgb]{0,0.502,0}{15\%$\downarrow$} 
    & \cellcolor[rgb]{1,1,0.741}\textbf{2.17} \textcolor[rgb]{0,0.502,0}{6\%$\downarrow$} \\

    Quadruped  
    & 2.36 & 0.68 
    & \cellcolor[rgb]{1,1,0.741}\textbf{2.01} \textcolor[rgb]{0,0.502,0}{15\%$\downarrow$} 
    & \cellcolor[rgb]{1,1,0.741}\textbf{0.63} \textcolor[rgb]{0,0.502,0}{7\%$\downarrow$} \\

    Pedestrian   
    & 2.87 & \textbf{1.20}
    & \cellcolor[rgb]{1,1,0.741}\textbf{2.85} \textcolor[rgb]{0,0.502,0}{1\%$\downarrow$} 
    & \cellcolor[rgb]{1,1,0.741}1.28 \textcolor[rgb]{1,0,0}{7\%$\uparrow$} \\

    Drone       
    & 1.20 & 1.17 
    & \cellcolor[rgb]{1,1,0.741}\textbf{1.17} \textcolor[rgb]{0,0.502,0}{3\%$\downarrow$} 
    & \cellcolor[rgb]{1,1,0.741}\textbf{1.14} \textcolor[rgb]{0,0.502,0}{3\%$\downarrow$} \\ 

    \hline

    Average     
    & 4.71 & 1.34 
    & \cellcolor[rgb]{1,1,0.741}\textbf{4.15} \textcolor[rgb]{0,0.502,0}{12\%$\downarrow$} 
    & \cellcolor[rgb]{1,1,0.741}\textbf{1.31} \textcolor[rgb]{0,0.502,0}{2\%$\downarrow$} \\

    \hline
    \end{tabular}%
}
\end{table*}

\subsection{Router Feature and Expert-Composition Analysis}
\label{app:router_analysis}

\begin{figure*}[h]  % [h] 表示这里(here)插入图片
    \centering      % 图片居中
    % 插入图片，支持 png/jpg/pdf/eps 格式
    \includegraphics[width=0.9\textwidth]{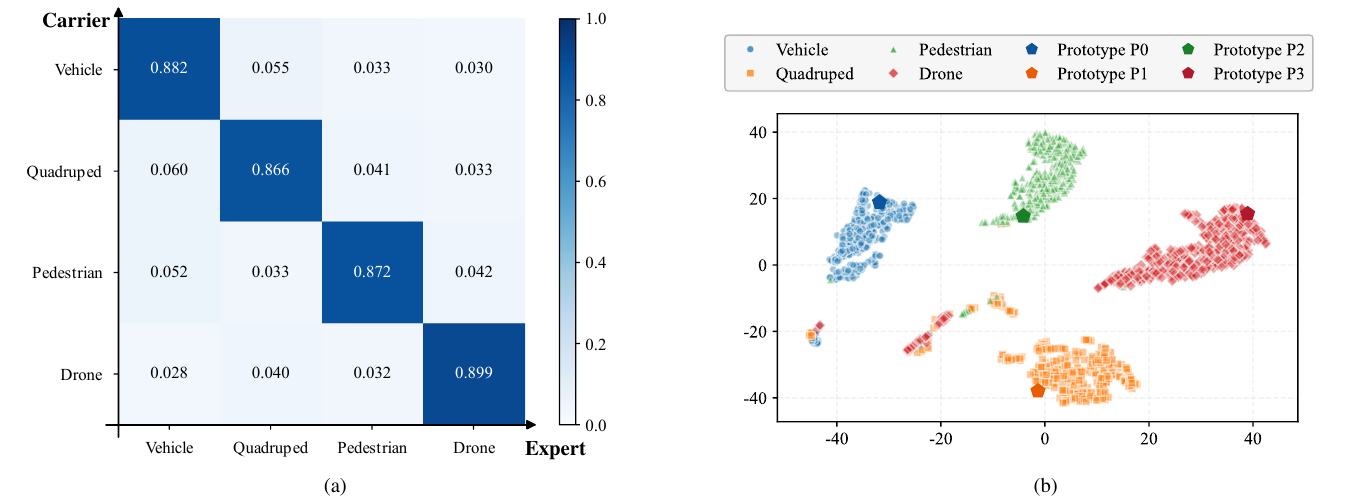}  
    \caption{\textbf{Router output weight and feature visualization.}
(a) Heatmap of routing-weight assignments between carriers and experts.
(b) t-SNE visualization of carrier features and learned prototypes.}  % 图片标题
     \label{fig:router_analysis}          % 图片标签（用于交叉引用）
\end{figure*}

We further analyze the behavior of the prototype router to better understand how MosaicIMU composes carrier-specific inertial features. 
As shown in Figure~\ref{fig:router_analysis}(a), the averaged routing matrix exhibits a clear diagonal-dominant structure. 
Each carrier assigns the largest routing weight to its corresponding expert, indicating that the learned prototypes preserve meaningful carrier semantics and that the experts develop carrier-specific inertial representations. 
Meanwhile, non-zero off-diagonal weights show that the router can also borrow complementary information from other carrier experts when forming the fused feature. 
This behavior is especially important for heterogeneous inertial data, where different platforms may share partial motion priors, vibration characteristics, or local velocity constraints despite having distinct overall dynamics.

Figure~\ref{fig:router_analysis}(b) visualizes the router feature embeddings together with the learned prototypes. 
Features from different carriers form distinguishable clusters, and the corresponding prototypes are located close to their associated feature regions, suggesting that the router learns a carrier-aware feature space. 
At the same time, the carrier clusters are not completely separated: local overlaps and neighboring regions can be observed among different carrier categories. 
This indicates that carrier dynamics are not purely discrete or mutually exclusive, but contain both carrier-specific and cross-carrier components. 
Therefore, imposing a rigid single-head prediction structure or a hard carrier-specific output partition may unnecessarily isolate useful shared inertial priors. 
In contrast, the proposed feature-level soft fusion preserves the dominant carrier-specific expert while allowing sample-adaptive feature sharing across experts, which provides a more flexible mechanism for multi-carrier inertial odometry.

\subsection{Ablation Studies of the Lightweight Offline Fine-tuning}
\label{app:ablation_finetune}
\begin{figure*}[h]  % [h] 表示这里(here)插入图片
    \centering      % 图片居中
    % 插入图片，支持 png/jpg/pdf/eps 格式
    \includegraphics[width=0.9\textwidth]{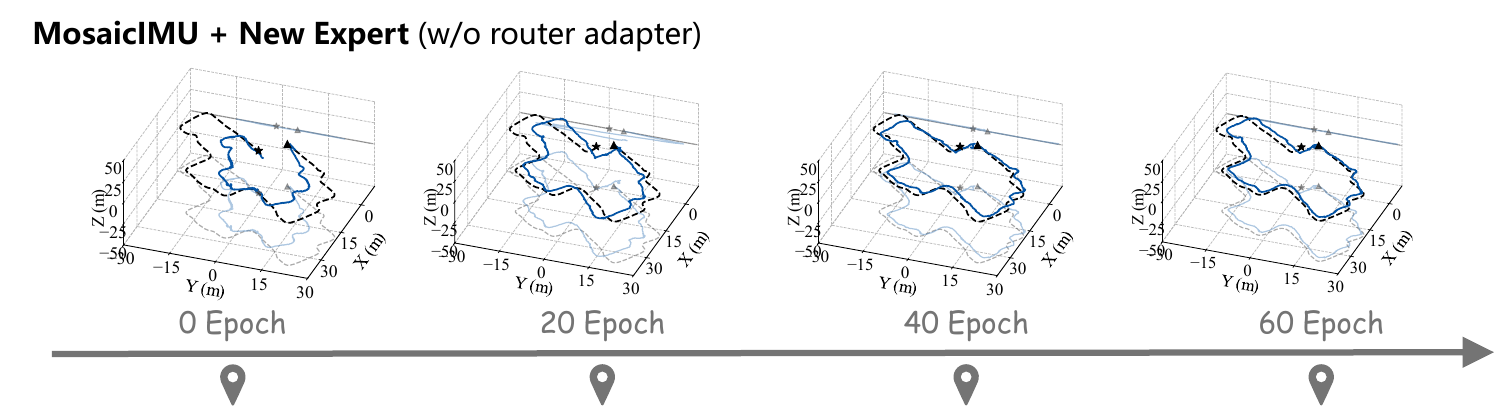}  
    \caption{\textbf{Effect of the router adapter on fine-tuning convergence.}
Trajectory estimates at different fine-tuning epochs without the router adapter.}  % 图片标题
     \label{fig:wo_router_adapter}          % 图片标签（用于交叉引用）
\end{figure*}

To further examine the role of the router adapter in the fine-tuning stage, we remove both the router adapter and the prior loss used to supervise the routing distribution. 
The results are reported in Figure~\ref{fig:wo_router_adapter} and Table~\ref{tab:wo_router_adapter}, while the corresponding results of MosaicIMU + New Expert are shown in Figure~\ref{fig:finetune}(a). 
The full method exhibits a much faster error convergence: it reaches the error level of the variant w/o adapter at 60 epochs using only about 20 epochs, and the error continues to decrease as the training proceeds. 
This demonstrates that carrier-prior guidance is effective for fine-tuning, as the corresponding carrier experts preserve motion knowledge that is beneficial for adapting to new data. 
However, due to the distribution mismatch between the features learned from the pretrained network and those of the new dataset, samples cannot be reliably assigned to the appropriate carrier experts. 
Introducing the router adapter alleviates this mismatch and enables the model to better exploit the prior knowledge stored in carrier-specific experts.

\begin{table*}[h]
\centering
\caption{\textbf{Ablation of the router adapter in lightweight fine-tuning.}
ATE and RTE-10s are reported at different epochs to compare convergence with and without the router adapter. Unit: m.}
\label{tab:wo_router_adapter}
\renewcommand{\arraystretch}{1.2}
\resizebox{\linewidth}{!}{
\begin{tabular}{c|cc|cc|cc|cc}
\hline
\multirow{2}{*}{Method} & \multicolumn{2}{c|}{0 Epoch} & \multicolumn{2}{c|}{20 Epoch} & \multicolumn{2}{c|}{40 Epoch} & \multicolumn{2}{c}{60 Epoch} \\
 & ATE $\downarrow$ & RTE-10s $\downarrow$ & ATE $\downarrow$ & RTE-10s $\downarrow$ & ATE $\downarrow$ & RTE-10s $\downarrow$ & ATE $\downarrow$ & RTE-10s $\downarrow$ \\
\hline
New Expert w/o router adapter & 9.85 & 5.36 & 8.88 & 2.44 & 3.20 & 1.55 & 2.64 & 1.12 \\
New Expert & 9.85 & 5.36 & \textbf{2.60} & \textbf{1.32} & \textbf{2.22} & \textbf{0.88} & \textbf{1.49} & \textbf{0.72} \\
\hline
\end{tabular}}
\end{table*}

\subsection{Details of Online Adaptation}
\label{app:online_details}

\begin{figure*}[h]  % [h] 表示这里(here)插入图片
    \centering      % 图片居中
    % 插入图片，支持 png/jpg/pdf/eps 格式
    \includegraphics[width=0.7\textwidth]{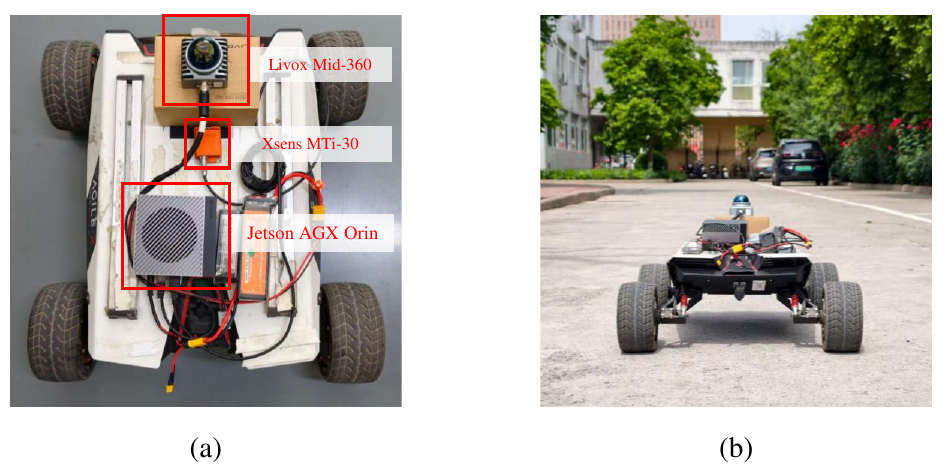}  
    \caption{\textbf{Online adaptation platform and test environment.}
    (a) On-board sensing and computing setup.
    (b) Four-wheeled vehicle operating on outdoor campus roads.}  % 图片标题
     \label{fig:online_configure}          % 图片标签（用于交叉引用）
\end{figure*}

We provide the implementation details of router-guided online adaptation. 
The experiment is conducted on a four-wheeled vehicle driving on outdoor campus roads. Figure~\ref{fig:online_configure} shows the on-board hardware configuration and the test environment, and Table~\ref{tab:online_configure} summarizes the main sensing and computing components. 
The trajectory in Figure~\ref{fig:online}(a) is used for online adaptation, while the trajectory in Figure~\ref{fig:online}(b) is a held-out sequence from the same target domain and is used only for evaluation. 
DLIO provides velocity supervision during online adaptation, but it is not used as an input to MosaicIMU during inference.

The online setting follows the lightweight adaptation strategy in Section~\ref{sec:incremental}: the pretrained base model is frozen, and only the router adapter and gated residual branch are updated. 
Since the target carrier is known to be a vehicle, we use the vehicle prior for router-guided adaptation. 
Incoming IMU windows are stored in an online buffer. 
At each update round, the calibrated router recomputes the mismatch score for the buffered windows, and only the Top-$M$ high-mismatch samples are used to update the lightweight branch. In this experiment, we set $M$ to 64.
This strategy keeps the update focused on informative samples while retaining past observations for replay.

The online update is deployed on a Jetson AGX Orin. 
Each epoch takes less than 10 seconds, after which newly collected windows are added to the buffer, and the selection-update procedure is repeated. 
In Figure~\ref{fig:online}(c), the vehicle finishes the adaptation trajectory after approximately 30 epochs. 
We then perform 10 additional epochs of buffer replay using the collected windows, which further stabilizes the adapted model.

\begin{table}[h]
\centering
\caption{\textbf{Specifications of the online adaptation platform.}
We summarize the main sensing, computing, and supervision components used in the online adaptation experiment.}
\label{tab:online_configure}
\renewcommand{\arraystretch}{1.2}
\resizebox{0.65\linewidth}{!}{
\begin{tabular}{c|c|c}
\hline
\textbf{Type} & \textbf{Model} & \textbf{Specification} \\
\hline
LiDAR & Livox Mid-360 & $\leq$2 cm, $1\sigma$, @10 m \\
IMU & Xsens MTi-30 & 18$^\circ$/h; 15 $\mu g$ \\
Edge computing platform & Jetson AGX Orin & 10.65 FP16 TFLOPS \\
\hline
\end{tabular}}
\end{table}

\end{document}